\documentclass[lettersize,journal]{IEEEtran}
\usepackage{amsmath,amsfonts}
\usepackage{algorithmic}
\usepackage{algorithm}
\usepackage{array}
\usepackage[caption=false,font=normalsize,labelfont=sf,textfont=sf]{subfig}
\usepackage{textcomp}
\usepackage{stfloats}
\usepackage{url}
\usepackage{verbatim}
\usepackage{graphicx}
\usepackage{cite}
\usepackage{booktabs}
\usepackage{multirow}
\usepackage{capt-of}
\usepackage[table]{xcolor}
\definecolor{spikeblue}{RGB}{225,244,252}
\colorlet{gain}{blue}
\definecolor{drop}{RGB}{255,15,15}
\definecolor{mydarkblue}{RGB}{0, 51, 153}   
\definecolor{wallcolor}{RGB}{174,199,232}
\definecolor{floorcolor}{RGB}{152,223,138}
\definecolor{cabinetcolor}{RGB}{31,119,180}
\definecolor{bedcolor}{RGB}{255,187,120}
\definecolor{chaircolor}{RGB}{188,189,34}
\definecolor{sofacolor}{RGB}{140,86,75}
\definecolor{tablecolor}{RGB}{255,152,150}
\definecolor{doorcolor}{RGB}{214,39,40}
\definecolor{windowcolor}{RGB}{197,176,213}
\definecolor{bookshelfcolor}{RGB}{148,103,189}
\definecolor{picturecolor}{RGB}{196,156,148}
\definecolor{countercolor}{RGB}{23,190,207}
\definecolor{deskcolor}{RGB}{247,182,210}
\definecolor{curtaincolor}{RGB}{219,219,141}
\definecolor{refrigeratorcolor}{RGB}{255,127,14}
\definecolor{showercolor}{RGB}{158,218,229}
\definecolor{toiletcolor}{RGB}{44,160,44}
\definecolor{sinkcolor}{RGB}{112,128,144}
\definecolor{bathtubcolor}{RGB}{227,119,194}
\definecolor{othercolor}{RGB}{82,84,163}
\definecolor{deltaRed}{RGB}{255,29,139}
\definecolor{deltaGreen}{RGB}{93,208,87}
\definecolor{resultgray}{RGB}{150,150,150}
\definecolor{s3disceilingcolor}{RGB}{0,255,0}
\definecolor{s3disfloorcolor}{RGB}{0,0,255}
\definecolor{s3diswallcolor}{RGB}{136,205,234}
\definecolor{s3disbeamcolor}{RGB}{255,255,0}
\definecolor{s3discolumncolor}{RGB}{218,116,218}
\definecolor{s3diswindowcolor}{RGB}{0,0,0}
\definecolor{s3disdoorcolor}{RGB}{162,147,126}
\definecolor{s3distablecolor}{RGB}{126,42,42}
\definecolor{s3dischaircolor}{RGB}{255,0,0}
\definecolor{s3dissofacolor}{RGB}{155,64,64}
\definecolor{s3disbookcasecolor}{RGB}{27,190,105}
\definecolor{s3disboardcolor}{RGB}{127,139,135}
\definecolor{s3discluttercolor}{RGB}{68,68,68}
\definecolor{ultralightgray}{RGB}{248, 248, 248}
\usepackage{makecell}
\usepackage[
    colorlinks=true,
    linkcolor=blue,
    citecolor=blue,
    urlcolor=blue
]{hyperref}
\definecolor{circlePurple}{RGB}{145, 92, 182}
\definecolor{circleRed}{HTML}{A50026}
\definecolor{circleOrange}{HTML}{FDBF11}
\newcommand{\best}[1]{\textbf{#1}}
\newcommand{\second}[1]{\underline{#1}}
\newcommand{\other}[1]{\textcolor{resultgray}{#1}}
\newcommand{\classlabel}[1]{%
    \rotatebox{90}{#1}%
}

\newcommand{\classcolor}[1]{%
    \textcolor{#1}{\rule{1.6ex}{1.6ex}}%
}

\begin{document}

\title{Parameter-Efficient Fine-Tuning for Spiking Point Cloud Models}

\author{Zihao Guo, Jihua Zhu,~\IEEEmembership{Senior
Member,~IEEE}, Yiding Sun, Dongxu Zhang, Hainan Luo,\\ Kai Tang, Shuaikang Zhu, Lin Chen, Jian Liu, and Ajmal Saeed Mian,~\IEEEmembership{Senior Member,~IEEE}
\thanks{This work was supported in part by the Natural Science Basis Research Plan in Shaanxi Province of China (No. 2025JC-JCQN
091), Technology Innovation Leading Program of Shaanxi (Program No.2024QY-SZX-23) and the Open Research Fund of Anhui Province Key Laboratory of Machine Vision Detection and Perception.}
\thanks{Zihao Guo, Jihua Zhu, Yiding Sun,  Dongxu Zhang, Shuaikang Zhu and
Lin Chen are with the School of Software Engineering, Xi’an Jiaotong
University, Xi’an 710048, China; Zihao Guo and Jihua Zhu are also with
the State Key Laboratory of Human-Machine Hybrid Augmented Intelligence,
Xi’an Jiaotong University, Xi’an 710048, China (e-mail: zhujh@xjtu.edu.cn)
(Co-corresponding author: Jihua Zhu and Hainan Luo).}
\thanks{Hainan Luo is with Wuhu HIT Robot Technology Research Institute Co., Ltd., Wuhu, China.}
\thanks{Kai Tang is with the School of Computing and Data Science, Nanyang Technological University, Singapore 639798 (e-mail: kai017@e.ntu.edu.sg).}
\thanks{Jian Liu is with the Department of Mechanical Engineering, College of Design and Engineering, National University of Singapore.}
\thanks{Ajmal Saeed Mian is with the Department of Computer Science, University of Western Australia, Perth, Australia.}
}

\markboth{Journal of \LaTeX\ Class Files,~Vol.~14, No.~8, August~2021}%
{Shell \MakeLowercase{\textit{et al.}}: A Sample Article Using IEEEtran.cls for IEEE Journals}


\maketitle

\begin{abstract}
Spiking Neural Networks (SNNs) offer energy-efficient 
solutions for point cloud analysis on resource-constrained 
devices through event-driven computation. However, existing 
pre-trained spiking point cloud models rely on full fine-tuning 
for downstream task adaptation, incurring substantial parameter 
and storage overhead. Furthermore, binary spike propagation 
suppresses task-relevant sub-threshold information. 
To address these issues, we propose SpikePEFT, 
the first parameter-efficient fine-tuning framework 
for spiking point cloud models. Specifically, 
Intrinsic Dynamics Tuner (IDT) adaptively modulates 
membrane decay and firing thresholds, enabling efficient 
neuron-intrinsic adaptation while keeping the pre-trained 
synaptic transformations frozen. Moreover, Silent-State 
Disambiguation Adaptation (SSDA) recovers task-relevant 
information from informative silent states, thereby providing 
richer evidence 
for downstream adaptation. Extensive experiments across multiple 
benchmarks demonstrate the effectiveness and efficiency 
of SpikePEFT. In particular, our SpikePEFT achieves 92.4\% 
accuracy on ModelNet40 and 85.6\% on the most 
challenging split ScanObjectNN(PB\_T50\_RS) while 
updating only about 5\% of the trainable 
parameters and preserving the energy efficiency of SNNs. 
For 3D dense prediction, SpikePEFT still remains competitive 
with full fine-tuning across diverse indoor and outdoor datasets. 
This work provides a promising step toward parameter-efficient 
adaptation of neuromorphic vision models.
\end{abstract}


\section{INTRODUCTION}

\IEEEPARstart{A}{rtificial} Neural Networks (ANNs) have achieved remarkable 
progress in point cloud 
analysis\cite{qi2017pointnet,pang2022pointmae,yu2022point,li2025pointdico}, 
yet such performance is typically accompanied by considerable 
computational and energy overhead. Energy inefficiency poses a 
major barrier to deploying advanced 3D perception systems on 
resource-constrained edge devices, such as drones, mobile robots, 
and AR/VR headsets. Real-world deployment scenarios demand reliable real-time processing under stringent power constraints. 
Consequently, the development of high-performance yet energy-efficient 
models for point cloud analysis is emerging as a critical and formidable 
research frontier, essential for advancing downstream applications.

Against this backdrop, bio-inspired Spiking Neural Networks (SNNs) provide a compelling alternative by leveraging event-driven computation and intrinsic spatio-temporal dynamics. Spiking PointNet\cite{ren2023spiking}, a pioneering spike-based point cloud method, extends PointNet with spiking neurons to perform event-driven feature extraction from unordered point sets. Following this line of research, recent studies have explored more effective spike encoding\cite{wu2025spiking,he2026hybrid}, local geometric modeling\cite{qiu2025efficient,dang2026primary}, and spatio-temporal feature extraction\cite{wu2025efficient} to improve the representation capability of SNNs for point clouds.

\begin{figure}[!t]
\centering
\includegraphics[width=1\columnwidth]{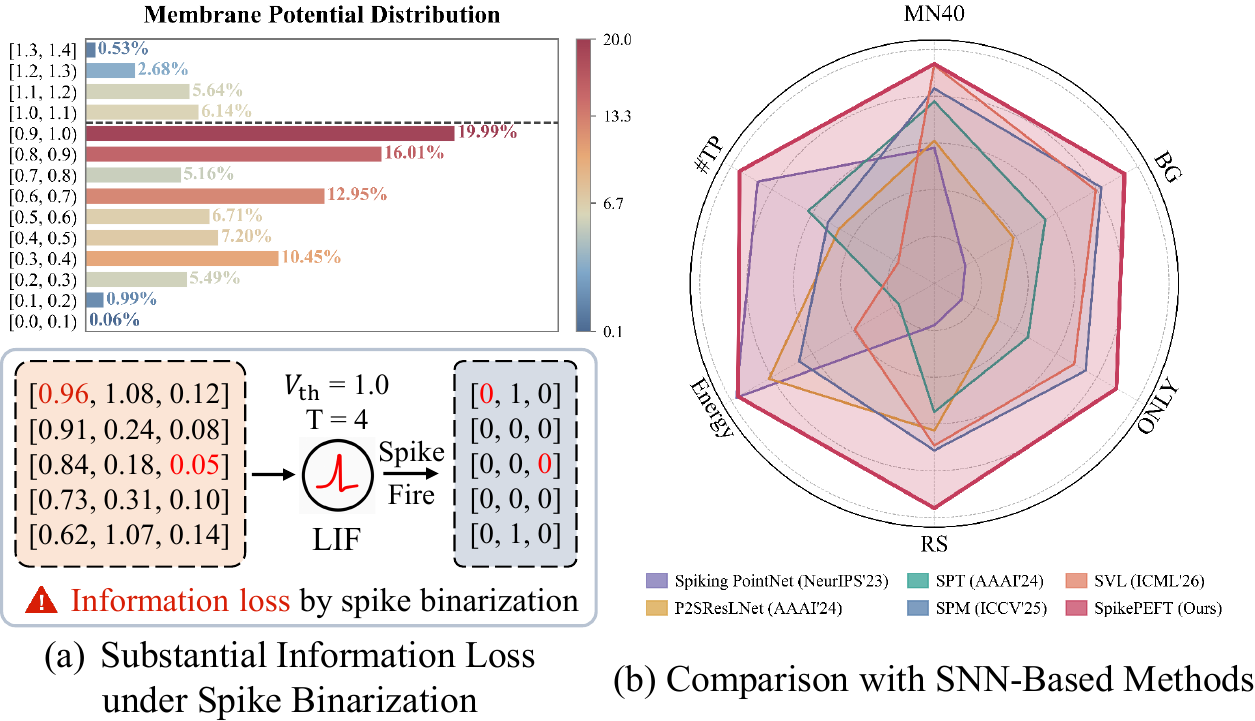} 
\caption{
    (a) Membrane potential distribution across the spiking blocks of the pre-trained SPM backbone on ModelNet40 with a time step of $T=4$.
    (b) Comparison with other representative SNN-based point cloud methods in terms of energy consumption (Energy), trainable parameters (\#TP), and overall accuracy on ModelNet40 (MN40) and three ScanObjectNN variants: OBJ\_BG (BG), OBJ\_ONLY (ONLY), and PB\_T50\_RS (RS). A larger radius indicates better performance, with Energy and \#TP inversely normalized.
}
\label{fig1}
\end{figure}

Despite recent advancements in SNN-based point cloud processing, 
critical limitations persist: (1) Adapting existing models to 
downstream tasks still relies on full fine-tuning, incurring 
substantial parameter and storage overhead. (2) Directly applying ANN-oriented PEFT methods to SNN backbones may
introduce dense continuous-valued transformations and additional MAC
operations. Accordingly, parameter efficiency in SNNs should not be characterized
solely by fewer trainable parameters, but also by whether sparse
event-driven computation and its inherent energy-efficiency advantage
are preserved. (3) As illustrated in Figure~\ref{fig1}(a), binary spike propagation
may obscure potentially informative differences among sub-threshold
membrane states during downstream adaptation, thereby limiting the
adaptability of pre-trained representations.

To solve the above problems simultaneously, we present \textit{SpikePEFT}, the first parameter-efficient fine-tuning (PEFT) framework for spiking point cloud models. 
First, we propose an \textit{Intrinsic Dynamics Tuner (IDT)} for lightweight adaptation of neuron-intrinsic temporal dynamics. Under the frozen synaptic backbone, IDT applies bounded residual adjustments to membrane decay and firing thresholds, enabling task-specific regulation of temporal integration and firing sensitivity without modifying the pre-trained synaptic transformations.
Second, we introduce a \textit{Silent-State Disambiguation Adapter (SSDA)}, which reformulates SNN adaptation as pre-spike membrane-state modulation rather than post-spike feature tuning. By measuring the margins between membrane states and firing thresholds, SSDA distinguishes near-threshold silent states from truly inactive states and selectively modulates the former before spike firing. Accordingly, task-relevant sub-threshold information suppressed by binary spikes can be incorporated into subsequent spike-driven propagation.

With these components, SpikePEFT outperforms full fine-tuning 
by 0.9~pp, 1.3~pp, and 1.4~pp on the three 
ScanObjectNN\cite{uy-scanobjectnn-iccv19} 
variants with SPM\cite{wu2025efficient}, 
respectively, while updating only 
approximately 5\% of the trainable parameters. 
For 3D dense prediction tasks, SpikePEFT remains competitive with full fine-tuning across diverse and challenging benchmarks, an uncommon achievement for PEFT methods.  
Notably, by optimizing only lightweight task-specific modules,
SpikePEFT substantially reduces the trainable parameters 
and downstream storage cost while preserving the spike-driven 
and energy-efficient computation of the pre-trained SNN backbone, 
as illustrated in Figure~\ref{fig1} (b). The main contributions 
of this work are summarized as follows:

\begin{itemize}
\item We reveal the limitations of existing downstream 
adaptation strategies 
for spiking point cloud models. 
Accordingly, we propose SpikePEFT, the first parameter-efficient fine-tuning framework 
for spiking point cloud models.

\item Building upon our insights into neuron-intrinsic 
dynamics, we propose IDT, 
which performs bounded residual adaptation of membrane 
decay and firing threshold while 
preserving the pre-trained synaptic transformations.

\item Derived from the neurophysiological notion of 
sub-threshold membrane dynamics, SSDA is proposed to 
selectively modulate near-threshold silent states before spike 
firing, thereby exploiting task-relevant information.

\item Extensive experiments demonstrate the effectiveness and efficiency of our approach, paving the way for efficient downstream adaptation of spiking point cloud models.

\end{itemize}

\section{Related Work}

\textbf{Spiking Neural Networks.} Spiking Neural Networks (SNNs) perform event-driven computation through asynchronous spikes on neuromorphic hardware, such as the Tianjic architecture\cite{pei2019towards}, exhibiting inherent compatibility with the sparsity of point cloud data. Spiking PointNet\cite{ren2023spiking}, the first SNN-based framework for point cloud analysis, proposes a `trained-less but learning-more' paradigm built upon PointNet\cite{qi2017pointnet}. Subsequently, P2SResLNet\cite{wu2024point} integrates spiking neurons with point convolutions to construct a point-to-spike residual network. SPT\cite{wu2025spiking} develops queue-driven encoding for spiking point cloud Transformers, while SPM\cite{wu2025efficient} establishes the first spike-based pre-training paradigm based on Spike Mamba. Along this trajectory, 3DSMT\cite{he2026hybrid} combines spiking Transformer\cite{vaswani2017attention} and Mamba modules\cite{gu2023mamba} for joint local and global feature modeling. While pre-trained spiking point cloud models with full fine-tuning have demonstrated exceptional capabilities for various downstream 3D tasks, the high computational costs and potential dilution of pre-trained knowledge motivate our exploration of efficient fine-tuning strategies.

\noindent\textbf{Parameter-Efficient Fine-Tuning.} Parameter-Efficient Fine-Tuning (PEFT) aims to adopt a trainable module with a few parameters for fine-tuning. It has attracted considerable attention in both the natural language processing\cite{houlsby2019parameter,li2021prefix,hu2022lora} and computer vision\cite{jia2022visual,jie2023revisiting,SUN2026112800,zhang2026pointcot}. PEFT approaches in 3D point cloud mainly fall into prompt-based\cite{zha2023idpt,ai2025gaprompt}, adapter-based\cite{liang2024pointgst,zha2025pma} and reparameterization-based\cite{han2025most,wang2025pointlora,sun2026tri,sun2026align}. For example, IDPT\cite{zha2023idpt} employs DGCNN\cite{wang2019dgcnn} to generate instance-aware dynamic prompts to improve robustness. Meanwhile, DAPT\cite{zhou2024dapt} couples prompt generation with dynamic adapter scaling to allocate adaptation capacity according to token relevance and MoST\cite{han2025most} brings Monarch-based sparse reparameterization with local geometric priors. However, existing 3D PEFT approaches are tailored to ANN-based backbones and operate with dense continuous computation. In contrast, our SpikePEFT, the first parameter-efficient fine-tuning framework for spiking point cloud models, fills the underexplored gap in PEFT for SNN-based backbones while preserving spike sparsity and energy efficiency.

\begin{figure*}[!t]
\centering
\includegraphics[width=1\textwidth]{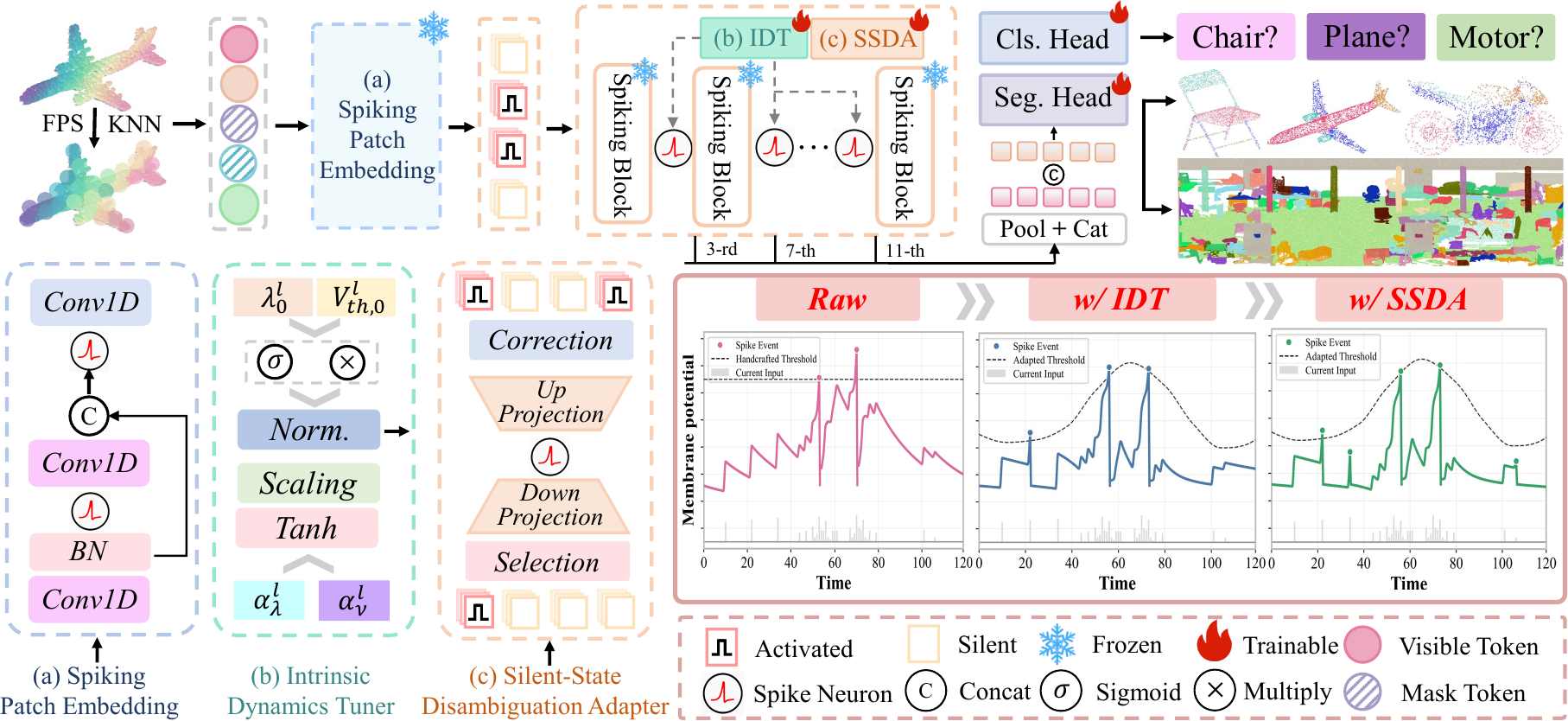}
\caption{Raw point clouds are sampled by Farthest Point Sampling (FPS), grouped into local patches via K-nearest neighbors (KNN), and processed by a frozen hierarchy of spiking blocks for downstream tasks. During training, the backbone is frozen and only newly added parameters are fine-tuned.
    (a) The Spiking Patch Embedding module first maps low-dimensional point coordinates into a high-dimensional spiking feature space, which serves as the input to the first spiking block.
    (b) IDT performs bounded residual tuning of membrane decay and firing thresholds while keeping synaptic transformations frozen.   
    (c) SSDA selects near-threshold silent states and processes only these activated positions through a lightweight spike-driven down-projection and up-projection to correct their pre-spike membrane potentials.
    The membrane trajectories illustrate the effects of IDT and SSDA on membrane potentials and spike firing.
}
\label{fig:pipeline}
\end{figure*}

\section{Method}



\subsection{Preliminaries}

\noindent\textbf{LIF Neuron.}
The Leaky Integrate-and-Fire (LIF) neuron describes a ``leaky-integrate-fire-reset'' process\cite{sedighi2024visual}. Given time step $t$, the LIF neuron is formulated as
\begin{equation}
H_t=f(V_{t-1},X_t),
\end{equation}
\begin{equation}
S_t=\Theta(H_t-V_{\rm th}),
\end{equation}
\begin{equation}
V_t=H_t\odot(1-S_t)+V_r\odot S_t,
\end{equation}
where $H_t$ and $V_t$ denote the membrane potentials after neuronal dynamics and 
after spike triggering, respectively. $X_t$ is the input, $V_{\rm th}$ 
is the firing threshold, $V_r$ is the reset potential, and $\Theta(\cdot)$ 
denotes the Heaviside function. 
Eq. (1) models the leaky integration process, 
where $f(\cdot)$ governs decay and input accumulation. 
$\odot$ denotes element-wise multiplication. We adopt the LIF neuron throughout the following sections.

\subsection{Intrinsic Dynamics Tuner}

The membrane decay and firing threshold jointly determine the
temporal behavior of an LIF neuron. The membrane decay controls the
retention and accumulation of historical membrane states, whereas the firing threshold
determines the sensitivity of spike generation. Therefore, instead of modifying the pre-trained synaptic transformations, the proposed IDT performs neuron-intrinsic adaptation through a lightweight task-specific parameterization.

For the $l$-th spiking block, let $\lambda_0^l$ and
$V_{{\rm th},0}^l$ denote the pre-trained membrane decay and firing
threshold, respectively. IDT introduces two trainable channel-wise
parameters
$a_{\lambda}^l,a_v^l\in\mathbb{R}^{d_l}$ and converts them into
bounded residuals:
\begin{equation}
\Delta_{\lambda}^l
=
\epsilon_{\lambda}\tanh(a_{\lambda}^l),
\qquad
\Delta_v^l
=
\epsilon_v\tanh(a_v^l),
\label{eq:idt_residual}
\end{equation}
where $\epsilon_{\lambda}$ and $\epsilon_v$ specify the allowable magnitudes of the residual adjustments. The adapted membrane decay and firing threshold
are then defined as:
\begin{equation}
\hat{\lambda}^l
=
\sigma\bigl(\operatorname{logit}(\lambda_0^l)+\Delta_{\lambda}^l\bigr),
\quad
\hat{V}_{\rm th}^l
=
V_{{\rm th},0}^l\odot \exp(\Delta_v^l),
\label{eq:idt_dynamics}
\end{equation}
where $\sigma(\cdot)$ denotes the sigmoid function. The logit-space adaptation
guarantees $0<\hat{\lambda}^l<1$, while the multiplicative
parameterization preserves $\hat{V}_{\rm th}^l>0$. Accordingly, IDT
adjusts the neuronal dynamics without producing invalid decay factors
or firing thresholds.

The adapted dynamics are incorporated into the membrane integration
and spike generation of the $l$-th spiking block:

\begin{equation}
\bar{H}_t^l
=
\hat{\lambda}^l
\odot V_{t-1}^l
+
W^lS_t^{l-1},
\label{eq:idt_membrane}
\end{equation}
\begin{equation}
S_t^{l,0}
=
\Theta\!\left(
\bar{H}_t^l-\hat{V}_{\rm th}^l
\right),
\label{eq:idt_spike}
\end{equation}
where $W^l$ is the frozen pre-trained synaptic transformation,
$\bar{H}_t^l$ denotes the adapted pre-spike membrane potential, and
$S_t^{l,0}$ is the preliminary spike output subsequently 
processed by SSDA. For multi-step backbones, 
both parameters are adapted to regulate temporal integration 
and firing sensitivity, whereas only $\hat{V}_{\rm th}^l$ 
is adapted for single-step backbones without historical 
membrane states. Since $W^l$ remains frozen, IDT modifies 
neuron-intrinsic dynamics and spike-generation behavior 
without modifying the pre-trained synaptic 
transformations.

\subsection{Silent-State Disambiguation Adapter}

Although IDT adapts temporal integration and firing sensitivity,
the information degradation induced by binary spike discretization
cannot be fully mitigated during downstream adaptation. Notably,
all sub-threshold membrane states are collapsed into the same
zero-valued output, rendering near-threshold neurons carrying
potentially task-relevant evidence indistinguishable from truly
inactive neurons. To resolve this silent-state ambiguity without
introducing a dense continuous adaptation pathway, we propose SSDA, which converts
near-threshold silent states into sparse binary events and performs
task-specific adaptation only at the activated positions.

For the $l$-th spiking block at time step $t$, IDT provides the
adapted pre-spike membrane potential
$\bar{H}_t^l\in\mathbb{R}^{N_l\times d_l}$ and the adapted firing
threshold $\hat{V}_{\rm th}^l\in\mathbb{R}^{d_l}$, where $N_l$ and
$d_l$ denote the number of spatial tokens and feature channels,
respectively. We employ $n\in\{1,\ldots,N_l\}$ and
$c\in\{1,\ldots,d_l\}$ to index the token and channel dimensions. The preliminary spike
$S_t^{l,0}\in\{0,1\}^{N_l\times d_l}$ identifies neurons that have
already crossed the adapted firing threshold.

To distinguish informative silent states, SSDA introduces a
learnable channel-wise interval:
\begin{equation}
\delta^l
=
\delta_{\max}^l
\odot
\sigma(a_{\delta}^l),
\qquad
a_{\delta}^l\in\mathbb{R}^{d_l},
\label{eq:ssda_interval}
\end{equation}
where $\sigma(\cdot)$ denotes the sigmoid function and $\odot$
denotes element-wise multiplication. $\delta_{\max}^l$ is a fixed
channel-wise upper bound for stable adaptation, while $a_{\delta}^l$ is trainable and
initialized to zero, yielding
$\delta^l=0.5\delta_{\max}^l$ at initialization.

Based on this interval, the near-threshold silent event is defined as:
\begin{equation}
Q_t^l
=
\left(1-S_t^{l,0}\right)
\odot
\Theta\!\left[
\bar{H}_t^l-
\left(
\hat{V}_{\rm th}^l-\delta^l
\right)
\right],
\label{eq:ssda_event}
\end{equation}
where $Q_t^l\in\{0,1\}^{N_l\times d_l}$ encodes the
near-threshold silent states. Therefore, already fired neurons and silent neurons far from the
firing threshold are excluded, while only near-threshold silent
states activate the adaptation pathway.

For the $n$-th token, we collect the channel indices associated with
active near-threshold events into
\begin{equation}
\Omega_{t,n}^l
=
\left\{
c\in\{1,\ldots,d_l\}
\mid
Q_{t,n,c}^l=1
\right\},
\label{eq:ssda_active_set}
\end{equation}
where $\Omega_{t,n}^l$ denotes the active near-threshold channel set. To model task-specific dependencies among the selected events, SSDA
employs a lightweight event-driven bottleneck. Let
$A^l\in\mathbb{R}^{r\times d_l}$ denote the trainable
down-projection matrix, where $r$ is the bottleneck dimension and
$r\ll d_l$. The resulting bottleneck current is denoted by
$U_t^l\in\mathbb{R}^{N_l\times r}$. Since $Q_t^l$ is binary, the
current for the $n$-th token is computed by accumulating only the
columns associated with active events:
\begin{equation}
U_{t,n,:}^l
=
\sum_{c\in\Omega_{t,n}^l}
A_{:,c}^l,
\label{eq:ssda_sparse_down}
\end{equation}
where $A_{:,c}^l\in\mathbb{R}^{r}$ denotes the $c$-th column of
$A^l$, while the colon denotes all entries along the corresponding
dimension. Similarly, $U_{t,n,:}^l$ denotes the complete
$r$-dimensional bottleneck current of the $n$-th token. This
operation accumulates only the weights indexed by active events and
avoids multiplication with continuous membrane activations.

The bottleneck current is subsequently converted into a binary spike:
\begin{equation}
Z_t^l
=
\Theta\!\left(
U_t^l-\vartheta_a
\right),
\label{eq:ssda_bottleneck_spike}
\end{equation}
where $\vartheta_a$ is a fixed firing threshold shared by the
bottleneck units. The resulting bottleneck spike satisfies
$Z_t^l\in\{0,1\}^{N_l\times r}$. We employ
$j\in\{1,\ldots,r\}$ to index the bottleneck channels and define
their active set as:
\begin{equation}
\Gamma_{t,n}^l
=
\left\{
j\in\{1,\ldots,r\}
\mid
Z_{t,n,j}^l=1
\right\},
\label{eq:ssda_bottleneck_set}
\end{equation}
where $\Gamma_{t,n}^l$ contains the indices of the active bottleneck
spikes for the $n$-th token.

Let $B^l\in\mathbb{R}^{d_l\times r}$ denote the trainable
up-projection matrix, and let
$R_t^l\in\mathbb{R}^{N_l\times d_l}$ denote the resulting
task-specific membrane correction. At an active near-threshold
position, the correction is computed as:
\begin{equation}
R_{t,n,c}^l
=
\sum_{j\in\Gamma_{t,n}^l}
B_{c,j}^l,
\qquad
c\in\Omega_{t,n}^l,
\label{eq:ssda_sparse_up}
\end{equation}
where $B_{c,j}^l$ denotes the element in the $c$-th row and $j$-th
column of $B^l$. Since $Z_t^l$ is binary, the up-projection is
implemented by accumulating only the weights associated with active
bottleneck spikes. Moreover, no correction is computed for channels
outside $\Omega_{t,n}^l$.

The task-specific correction is evaluated and applied exclusively
at the identified near-threshold positions:
\begin{equation}
\tilde{H}_{t,n,c}^l
=
\begin{cases}
\bar{H}_{t,n,c}^l + R_{t,n,c}^l,
&
c\in\Omega_{t,n}^l,
\\
\bar{H}_{t,n,c}^l,
&
c\notin\Omega_{t,n}^l.
\end{cases}
\label{eq:ssda_sparse_correction}
\end{equation}
The up-projection matrix $B^l$ is initialized to zero, ensuring
that $\tilde{H}_t^l=\bar{H}_t^l$ at initialization and thus avoiding abrupt perturbations to the pre-trained neuronal dynamics. 
Therefore, SSDA initially preserves the original membrane evolution and progressively learns task-specific membrane corrections during downstream adaptation. In
implementation, the correction is applied through indexed
scatter-add operations only over the activated positions specified
by $\Omega_{t,n}^l$, thereby maintaining the sparse computation of the spiking backbone. Finally, the spike output and reset
membrane potential are computed as:
\begin{equation}
S_t^l
=
\Theta\!\left(
\tilde{H}_t^l-\hat{V}_{\rm th}^l
\right),
\label{eq:ssda_spike}
\end{equation}
\begin{equation}
V_t^l
=
\tilde{H}_t^l\odot(1-S_t^l)
+
V_r^l\odot S_t^l,
\label{eq:ssda_reset}
\end{equation}
where $S_t^l\in\{0,1\}^{N_l\times d_l}$ denotes the final binary
spike propagated to the subsequent spiking block, and $V_r^l$
denotes the reset potential of the backbone neurons. 

During training, the equivalent tensor 
$U_t^l=Q_t^l(A^l)^\top$ and
$R_t^l=Q_t^l\odot[Z_t^l(B^l)^\top]$ are employed to enable
surrogate-gradient optimization. During inference, sparse weight
accumulation is performed over the nonzero event indices.

\begin{table*}[!t]
\caption{Classification results on ModelNet40\cite{wu20153d}, ModelNet10, and
three variants of ScanObjectNN\cite{uy-scanobjectnn-iccv19}. $\dagger$, $\ddagger$, and $\diamond$ denote supervised learning, self-supervised learning, and PEFT methods, respectively.}
\centering
\setlength{\tabcolsep}{3pt}
\renewcommand{\arraystretch}{0.95}
\small

\resizebox{\textwidth}{!}{
\begin{tabular}{cllccccccccc}
\toprule

\multirow{2}{*}{Type}
& \multicolumn{1}{l}{\multirow[c]{2}{*}{Methods}}
& \multirow{2}{*}{Reference}
& \multirow{2}{*}{Input}
& \multirow{2}{*}{\#TP (M)\ $\downarrow$}
& \multirow{2}{*}{$T$}
& \multirow{2}{*}{Energy (mJ)\ $\downarrow$}
& \multirow{2}{*}{ModelNet10\ $\uparrow$}
& \multirow{2}{*}{ModelNet40\ $\uparrow$}
& \multicolumn{3}{c}{ScanObjectNN\ $\uparrow$}
\\

\cmidrule(lr){10-12}

& & & & & & & & 
& OBJ\_BG
& OBJ\_ONLY
& PB\_T50\_RS
\\

\midrule

\multirow[c]{8}{*}{ANN}

& PointNet$^{\dagger}$\cite{qi2017pointnet}
& CVPR2017
& Point
& 1.7
& --
& 2.0
& 93.0
& 90.7
& 82.3
& 84.3
& 77.9
\\

& Point-BERT$^{\ddagger}$\cite{yu2022point}
& CVPR2022
& Point
& 22.1
& --
& 22.1
& 93.0
& 93.2
& 87.4
& 88.1
& 83.1
\\

& Point-MAE$^{\ddagger}$\cite{pang2022pointmae}
& ECCV2022
& Point
& 22.1
& --
& \ \ 22.1$^{*}$
& 93.1
& 93.2
& 90.0
& 88.3
& 85.2
\\

& IDPT$^{\diamond}$\cite{zha2023idpt}
& ICCV2023
& Point
& 1.7
& --
& \ \ 29.6$^{*}$
& 93.9
& 93.3
& 91.2
& 90.0
& 84.5
\\

& PointMamba$^{\ddagger}$\cite{liang2024pointmamba}
& NeurIPS2024
& Point
& 12.3
& --
& 18.9
& 93.3$^{*}$
& 92.4
& 90.2
& 89.6
& 85.4
\\

& SI-Mamba$^{\ddagger}$\cite{bahri2025spectral}
& CVPR2025
& Point
& 12.3
& --
& \ \ 16.6$^{*}$
& 93.5$^{*}$
& 92.7
& 92.3
& 91.4
& 87.3
\\

& PointLoRA$^{\diamond}$\cite{wang2025pointlora}
& CVPR2025
& Point
& 0.8
& --
& \ \ 33.5$^{*}$
& 94.1$^{*}$
& 93.3
& 90.7
& 89.3
& 85.5
\\

& PointATA$^{\diamond}$\cite{sun2026align}
& TMM2026
& Point
& 1.1
& --
& \ \ 24.1
& 94.4$^{*}$
& 92.7
& 90.9
& 89.7
& 85.4
\\

\midrule

\multirow[c]{10}{*}{SNN}

& Spike PointNet$^{\dagger}$\cite{lan2023efficient}
& ICCV2023
& Point
& 3.5
& 16
& 0.1
& 92.4
& 88.6
& --
& --
& 69.2
\\

& SpikingPointNet$^{\dagger}$\cite{ren2023spiking}
& NeurIPS2023
& Point
& 3.5
& 4
& 0.4
& 93.3
& 88.2
& 72.2
& 76.4
& 64.1
\\

& P2SResLNet-B$^{\dagger}$\cite{wu2024point}
& AAAI2024
& Point
& 14.3
& 1
& 3.0
& 93.5
& \ \ 88.3$^{*}$
& 78.6
& 80.2
& 74.5
\\

& E-3DSNN-L$^{\dagger}$\cite{qiu2025efficient}
& AAAI2025
& Voxel
& 17.7
& 1
& 0.3
& 93.4$^{*}$
& 91.2
& \ \ 83.1$^{*}$
& \ \ 84.7$^{*}$
& 80.2
\\

& SPT$^{\dagger}$\cite{wu2025spiking}
& AAAI2025
& Point
& 10.2
& 4
& 13.3
& 93.7$^{*}$
& 91.4
& 82.8
& 83.4
& 78.0
\\

& SDT$^{\dagger}$\cite{lu2026spiking}
& ICLR2026
& Point
& 2.3
& 4
& 2.1
& --
& 92.5
& --
& --
& \textbf{86.2}
\\

\cmidrule(l){2-12}

& SPM$^{\ddagger}$\cite{wu2025efficient}
& ICCV2025
& Point
& 12.8\ (100\%)
& 4
& 5.4
& 94.1
& 92.3
& \underline{90.2}
& \underline{89.5}
& 84.2
\\

& \cellcolor{spikeblue}\textbf{w/ SpikePEFT (Ours)}$^{\diamond}$
& \multicolumn{1}{c}{\cellcolor{spikeblue}--}
& \cellcolor{spikeblue}Point
& \cellcolor{spikeblue}0.7\ (5.2\%)
& \cellcolor{spikeblue}4
& \cellcolor{spikeblue}6.2
& \cellcolor{spikeblue}94.7\ \textcolor{gain}{(+\ 0.6)}
& \cellcolor{spikeblue}92.4\ \textcolor{gain}{(+\ 0.1)}
& \cellcolor{spikeblue}\textbf{91.1}\ \textcolor{gain}{(+\ 0.9)}
& \cellcolor{spikeblue}\textbf{90.8}\ \textcolor{gain}{(+\ 1.3)}
& \cellcolor{spikeblue}\underline{85.6}\ \textcolor{gain}{(+\ 1.4)}
\\

\cmidrule(l){2-12}

& E-3DSNN-L + SVL$^{\ddagger}$\cite{qiu2025svl}
& ICML2026
& Voxel
& 17.7\ (100\%)
& 1
& 0.3
& 94.5$^{*}$
& \textbf{93.7}
& 84.8$^{*}$
& 85.2$^{*}$
& 83.0
\\

& \cellcolor{spikeblue}\textbf{w/ SpikePEFT (Ours)}$^{\diamond}$
& \multicolumn{1}{c}{\cellcolor{spikeblue}--}
& \cellcolor{spikeblue}Voxel
& \cellcolor{spikeblue}0.8\ (4.6\%)
& \cellcolor{spikeblue}1
& \cellcolor{spikeblue}0.5
& \cellcolor{spikeblue}95.0\ \textcolor{gain}{(+\ 0.5)}
& \cellcolor{spikeblue}\underline{93.6}\ \textcolor{drop}{(-\ 0.1)}
& \cellcolor{spikeblue}85.7\ \textcolor{gain}{(+\ 0.9)}
& \cellcolor{spikeblue}85.9\ \textcolor{gain}{(+\ 0.7)}
& \cellcolor{spikeblue}84.1\ \textcolor{gain}{(+\ 1.1)}
\\

\bottomrule
\end{tabular}
}

\label{tab:classification}
\vspace{-6pt}
\end{table*}

\begin{table*}[!t]
\centering
\begin{minipage}[t]{0.49\textwidth}
\captionof{table}{Part segmentation results on the
ShapeNetPart\cite{shapenetpart}.}
\centering
\setlength{\tabcolsep}{1pt}
\renewcommand{\arraystretch}{1.05}
\small

\resizebox{\linewidth}{!}{
\begin{tabular}{clccccc}
\toprule
Type
& \multicolumn{1}{l}{Methods}
& \#TP
& \ Energy
& allAcc
& Cls.\ mIoU
& Ins.\ mIoU
\\
\midrule

\multirow[c]{5}{*}{ANN}
& Point-BERT\cite{yu2022point}
& 27.1
& \ 72.0$^{*}$
& \ 89.2$^{*}$
& 84.1
& 85.6
\\

& Point-MAE\cite{pang2022pointmae}
& 27.1
& \ 71.7$^{*}$
& \ 89.5$^{*}$
& 84.1
& 86.1
\\

& ReCon\cite{qi2023recon}
& 48.5
& \ 75.9$^{*}$
& \ 89.8$^{*}$
& 84.5
& 86.4
\\

& PointMamba\cite{liang2024pointmamba}
& 17.4
& \ 68.1$^{*}$
& 88.7
& 82.6
& 85.3
\\

& PointATA\cite{sun2026align}
& 7.3
& 70.6
& 88.6
& 84.8
& 85.7
\\

\midrule

\multirow[c]{7}{*}{SNN}

& E-3DSNN-L\cite{qiu2025efficient}
& 20.1
& \ 1.1$^{*}$
& \ 87.5$^{*}$
& 81.7
& 83.8
\\

& SPT\cite{wu2025spiking}
& 19.5
& \ 25.6$^{*}$
& \ 87.0$^{*}$
& 81.3
& 82.9
\\

& SDT\cite{lu2026spiking}
& 4.6
& 4.7
& --
& \textbf{83.7}
& 85.1
\\

\cmidrule(l){2-7}

& SPM\cite{wu2025efficient}
& 18.3
& 12.3
& \ 88.4$^{*}$
& 82.3
& 84.8
\\

& \cellcolor{spikeblue}\textbf{w/ SpikePEFT (Ours)}
& \cellcolor{spikeblue}6.0
& \cellcolor{spikeblue}13.4
& \cellcolor{spikeblue}\textbf{89.1}\textcolor{gain}{(+\ 0.7)}
& \cellcolor{spikeblue}\underline{83.6}\textcolor{gain}{(+\ 1.3)}
& \cellcolor{spikeblue}\underline{85.2}\textcolor{gain}{(+\ 0.4)}
\\

\cmidrule(l){2-7}

& E-3DSNN-L + SVL\cite{qiu2025svl}
& 20.1
& \ 1.2$^{*}$
& \ 88.6$^{*}$
& 82.8
& 85.0
\\

& \cellcolor{spikeblue}\textbf{w/ SpikePEFT (Ours)}
& \cellcolor{spikeblue}6.2
& \cellcolor{spikeblue}\ 1.5$^{*}$
& \cellcolor{spikeblue}\underline{89.0}\textcolor{gain}{(+\ 0.4)}
& \cellcolor{spikeblue}\underline{83.6}\textcolor{gain}{(+\ 0.8)}
& \cellcolor{spikeblue}\textbf{85.7}\textcolor{gain}{(+\ 0.7)}
\\

\bottomrule
\end{tabular}
}
\label{tab:part_segmentation}
\end{minipage}
\hfill
\begin{minipage}[t]{0.49\textwidth}
\captionof{table}{
Semantic segmentation results on S3DIS Area 5\cite{Armeni_2016_CVPR}.
}
\centering
\setlength{\tabcolsep}{1pt}
\renewcommand{\arraystretch}{1.05}
\small

\resizebox{\linewidth}{!}{
\begin{tabular}{clccccc}
\toprule
Type
& \multicolumn{1}{l}{Methods}
& \#TP
& \ Energy
& mAcc
& allAcc
& mIoU
\\
\midrule

\multirow[c]{5}{*}{ANN}
& Point-BERT\cite{yu2022point}
& 27.0
& \ 119.7$^{*}$
& 69.7
& 88.5
& 60.5
\\

& Point-MAE\cite{pang2022pointmae}
& 27.0
& \ 120.2$^{*}$
& 69.9
& 88.8
& 60.8
\\

& ReCon\cite{qi2023recon}
& 48.5
& \ 126.2$^{*}$
& 69.3
& 88.3
& 60.4
\\

& PointMamba\cite{liang2024pointmamba}
& 17.4
& \ 115.0$^{*}$
& 68.2
& 87.6
& 58.4
\\

& PointATA\cite{sun2026align}
& 7.2
& 127.4
& 67.6
& 87.8
& 59.1
\\

\midrule

\multirow[c]{7}{*}{SNN}

& E-3DSNN-L\cite{qiu2025efficient}
& 20.1
& \ 1.9$^{*}$
& \ 64.3$^{*}$
& 86.9
& \ 60.2$^{*}$
\\

& SPT\cite{wu2025spiking}
& 19.5
& \ 38.2$^{*}$
& \ 67.8$^{*}$
& 88.1
& \ 62.3$^{*}$
\\

& SDT\cite{lu2026spiking}
& 10.7
& 7.3
& \textbf{76.8}
& \textbf{91.2}
& \textbf{69.6}
\\

\cmidrule(l){2-7}

& SPM\cite{wu2025efficient}
& 18.3
& 19.1$^{*}$
& 68.8
& \underline{88.6}
& 63.3
\\

& \cellcolor{spikeblue}\textbf{w/ SpikePEFT (Ours)}
& \cellcolor{spikeblue}6.0
& \cellcolor{spikeblue}20.4
& \cellcolor{spikeblue}\underline{69.2}\textcolor{gain}{(+\ 0.4)}
& \cellcolor{spikeblue}\underline{89.0}\textcolor{gain}{(+\ 0.4)}
& \cellcolor{spikeblue}\underline{64.0}\textcolor{gain}{(+\ 0.7)}
\\

\cmidrule(l){2-7}

& E-3DSNN-L + SVL\cite{qiu2025svl}
& 20.1
& 2.1$^{*}$
& 65.4
& 87.4
& 61.4
\\

& \cellcolor{spikeblue}\textbf{w/ SpikePEFT (Ours)}
& \cellcolor{spikeblue}6.1
& \cellcolor{spikeblue}2.4
& \cellcolor{spikeblue}65.6\textcolor{gain}{(+\ 0.2)}
& \cellcolor{spikeblue}88.2\textcolor{gain}{(+\ 0.8)}
& \cellcolor{spikeblue}61.4\textcolor{gain}{(+\ 0.0)}
\\

\bottomrule
\end{tabular}
}
\label{tab:semantic_segmentation}
\end{minipage}
\end{table*}

\section{Experiments}

We extensively evaluate SpikePEFT on multiple downstream tasks, including
object classification, part segmentation, and semantic segmentation.
We employ two  pre-trained and frozen spiking point cloud models, SPM and
E-3DSNN-L with SVL, as the backbones. For a fair comparison, we follow the default fine-tuning protocols of each backbone, including data preprocessing, task-specific
heads, and optimization settings. We implement SpikePEFT in PyTorch 2.7.1\cite{paszke2019pytorch} on $2\times$ NVIDIA A800-SXM4-80GB GPUs, with the spiking components developed using the SpikingJelly framework\cite{fang2023spikingjelly}. 

For evaluation, we report overall accuracy for object classification, class- and instance-level mIoU (Cls. mIoU and Ins. mIoU) together with overall accuracy (allAcc) for part segmentation, and mean accuracy (mAcc), allAcc, and mean IoU (mIoU) for semantic segmentation. We additionally report the number of trainable 
parameters (\#TP), time steps ($T$), and energy consumption to evaluate parameter and computational efficiency. 
$^{*}$ denotes results reproduced from the publicly available source code. Among the SNN-based methods, the best
two results are highlighted in \textbf{bold} and
\underline{underlined}.

\subsection{Object Classification}

\noindent\textbf{Synthetic Object Classification.}
The \textit{ModelNet10}\cite{wu20153d} is a compact dataset
comprising 4,899 3D models from 10 object categories. As a subset of \textit{ModelNet40}, it contains fewer categories
with more pronounced inter-class differences.
In comparison, \textit{ModelNet40}\cite{wu20153d}
comprises 12,311 3D CAD models across 40 categories. Due to the additional computational cost of the voting 
strategy\cite{liu2019relation}, 
we report overall 
accuracy without voting. 
As shown in Table~\ref{tab:classification}, 
compared to full fine-tuning, SpikePEFT substantially 
reduces downstream computational resource requirements 
while achieving comparable performance. 
Notably, this performance gain incurs only a 
practically negligible increase in energy consumption. 
Figure~\ref{fig:tsne} presents the t-SNE feature clustering results. 
SpikePEFT yields lower intra-class and higher inter-class distances,
indicating improved feature compactness and separability.

\noindent\textbf{Real-World Object Classification.}
\textit{ScanObjectNN}\cite{uy-scanobjectnn-iccv19} is a highly 
challenging 3D dataset covering $\sim$15K diverse real-world objects 
across 15 categories. These objects consist of indoor 
scene data obtained by scanning, often characterized by 
cluttered backgrounds and occlusion caused by other objects.
As shown in Table~\ref{tab:classification}, SpikePEFT 
surpasses full fine-tuning of SPM and E-3DSNN-L with 
SVL by 1.4~pp and 1.1~pp, respectively, on the challenging PB\_T50\_RS 
split, while training only approximately 5\% of the parameters.
This suggests that SpikePEFT reduces potentially destructive 
changes to pre-trained spiking representations while 
selectively enhancing task-relevant neuronal responses, leading to more robust 
recognition under challenging real-world perturbations.

\subsection{3D Dense Prediction Task} 
\begin{figure}[!t]
\centering
\includegraphics[width=1\columnwidth]{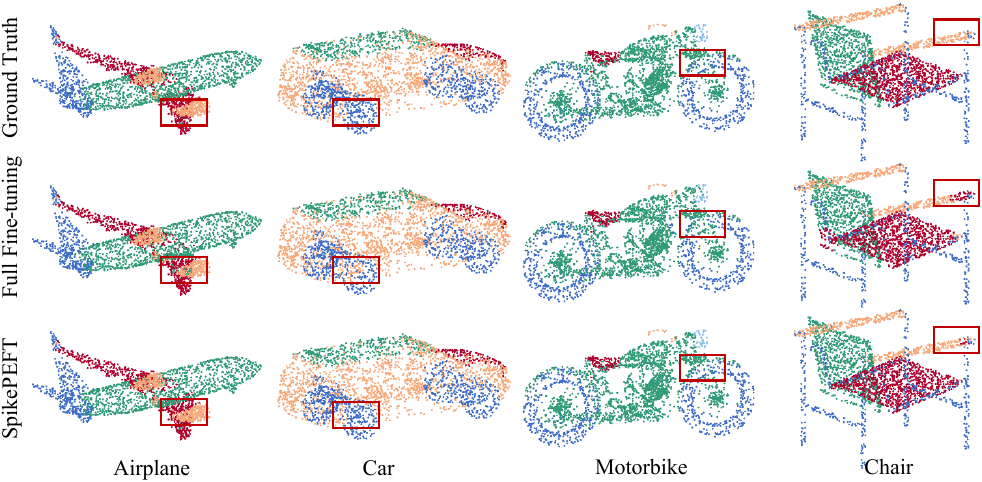}
\vspace{-14pt}
\caption{Qualitative analysis results for part segmentation on ShapeNetPart.}
\label{fig:part}
\vspace{-14pt}
\end{figure}

\noindent\textbf{Part Segmentation.} We conduct 
part segmentation experiments on \textit{ShapeNetPart}\cite{shapenetpart} dataset, which includes 16,881 
samples from 16 categories and 50 annotated part labels.
As shown in Table~\ref{tab:part_segmentation}, in this fine-grained scene 
understanding task, SpikePEFT still achieves the best
or second-best performance among SNN-based methods. 
As illustrated in Figure~\ref{fig:part},
by preserving richer task-relevant information, SpikePEFT produces
more accurate and coherent part boundaries than the baselines,
especially for small or structurally complex parts. 
Distinct from the classification task, the increase primarily comes from the
5.3M segmentation head, while SpikePEFT updates only
0.7M parameters (5.5\% of the backbone) in SPM.

\begin{table*}[!t]
\centering

\caption{Semantic segmentation results on
ScanNet V2\cite{dai2017scannet}.}
\label{tab:scannet}

\fontsize{4.2}{5.3}\selectfont
\setlength{\tabcolsep}{1pt}
\renewcommand{\arraystretch}{1.02}

\resizebox{\textwidth}{!}{%
\begin{tabular}{cl|ccc|*{20}{c}}
\toprule

\multirow[c]{2}{*}{%
  \raisebox{2.8ex}[0pt][0pt]{Type}%
}
&
\multirow[c]{2}{*}{%
  \raisebox{2.8ex}[0pt][0pt]{Method}%
}
&
\multicolumn{3}{c|}{%
  \raisebox{4.0ex}[0pt][0pt]{ScanNet V2\cite{dai2017scannet}}%
}
&
\classlabel{wall}
&
\classlabel{floor}
&
\classlabel{cabinet}
&
\classlabel{bed}
&
\classlabel{chair}
&
\classlabel{sofa}
&
\classlabel{table}
&
\classlabel{door}
&
\classlabel{window}
&
\classlabel{books.}
&
\classlabel{picture}
&
\classlabel{counter}
&
\classlabel{desk}
&
\classlabel{curtain}
&
\classlabel{refrig.}
&
\classlabel{shower}
&
\classlabel{toilet}
&
\classlabel{sink}
&
\classlabel{bathtub}
&
\classlabel{other}
\\[-2.8ex]

\noalign{%
  \global\advance\aboverulesep by -0.5ex
  \global\advance\belowrulesep by 0.5ex
}
\cmidrule{3-5}
\noalign{%
  \global\advance\aboverulesep by 0.5ex
  \global\advance\belowrulesep by -0.5ex
}

&
&
\raisebox{1ex}[0pt][0pt]{\#TP (M)}
&
\raisebox{1ex}[0pt][0pt]{mAcc}
&
\raisebox{1ex}[0pt][0pt]{mIoU}
&
\classcolor{wallcolor}
&
\classcolor{floorcolor}
&
\classcolor{cabinetcolor}
&
\classcolor{bedcolor}
&
\classcolor{chaircolor}
&
\classcolor{sofacolor}
&
\classcolor{tablecolor}
&
\classcolor{doorcolor}
&
\classcolor{windowcolor}
&
\classcolor{bookshelfcolor}
&
\classcolor{picturecolor}
&
\classcolor{countercolor}
&
\classcolor{deskcolor}
&
\classcolor{curtaincolor}
&
\classcolor{refrigeratorcolor}
&
\classcolor{showercolor}
&
\classcolor{toiletcolor}
&
\classcolor{sinkcolor}
&
\classcolor{bathtubcolor}
&
\classcolor{othercolor}
\\

\midrule


\multirow[c]{5}{*}{ANN}

& Point-MAE\cite{pang2022pointmae}
& 27.0
& 81.3
& 69.3
& 86.0
& 97.1
& 68.8
& 81.8
& 89.2
& 71.4
& 68.6
& 53.0
& 60.6
& 68.8
& 50.9
& 60.9
& 54.8
& 69.6
& 69.6
& 62.8
& 86.4
& 64.6
& 70.8
& 50.3
\\

& PointNeXt\cite{qian2022pointnext}
& 41.6
& 82.0
& 71.5
& 92.1
& 97.0
& 67.7
& 82.2
& 84.3
& 77.3
& 73.5
& 58.8
& 67.5
& 62.5
& 57.7
& 60.2
& 63.5
& 67.5
& 76.0
& 58.7
& 87.3
& 62.6
& 75.0
& 58.6
\\

& ReCon\cite{qi2023recon}
& 48.5
& 83.1
& 72.0
& 89.3
& 94.5
& 71.3
& 80.2
& 87.0
& 80.0
& 71.2
& 52.4
& 61.4
& 68.8
& 51.3
& 62.4
& 67.4
& 68.6
& 79.4
& 63.9
& 88.8
& 67.1
& 76.9
& 58.1
\\

& PointMamba\cite{liang2024pointmamba}
& 17.4
& 83.4
& 72.2
& 91.3
& 97.1
& 72.9
& 80.2
& 81.7
& 77.3
& 77.0
& 64.4
& 59.8
& 62.5
& 54.4
& 62.9
& 66.4
& 76.2
& 82.3
& 61.5
& 89.9
& 61.5
& 71.1
& 53.6
\\
& PointATA\cite{sun2026align}
& 7.2
& 83.5
& 72.6
& 90.8
& 97.3
& 72.4
& 81.6
& 85.1
& 77.0
& 74.2
& 60.7
& 64.1
& 65.4
& 55.9
& 63.5
& 64.8
& 71.8
& 78.4
& 62.9
& 89.1
& 66.1
& 74.0
& 56.9
\\

\midrule

\multirow[c]{7}{*}{SNN}

& P2SResLNet-B\cite{wu2024point}
& 23.1
& \other{81.7}
& \other{69.1}
& \other{87.8}
& \other{90.7}
& \other{69.1}
& \other{81.4}
& \other{77.0}
& \other{70.9}
& \other{71.2}
& \other{55.9}
& \other{60.8}
& \other{62.3}
& \other{55.5}
& \other{61.2}
& \second{60.4}
& \other{69.9}
& \other{76.3}
& \other{57.3}
& \other{87.2}
& \other{62.0}
& \second{73.1}
& \other{52.0}
\\

& E-3DSNN-L\cite{qiu2025efficient}
& 20.1
& \other{82.5}
& \other{69.7}
& \other{86.4}
& \other{91.9}
& \other{70.6}
& \other{78.5}
& \other{81.6}
& \other{73.9}
& \other{72.5}
& \other{58.0}
& \other{62.2}
& \second{66.1}
& \other{57.2}
& \other{57.7}
& \other{60.2}
& \other{64.5}
& \other{76.8}
& \other{63.3}
& \other{86.6}
& \other{62.4}
& \other{68.4}
& \other{55.2}
\\

& SPT\cite{wu2025spiking}
& 19.5
& \other{82.9}
& \other{70.4}
& \other{89.8}
& \other{96.2}
& \other{65.6}
& \other{80.3}
& \other{83.7}
& \second{76.2}
& \other{73.4}
& \other{58.3}
& \other{63.6}
& \other{65.8}
& \other{54.7}
& \other{61.5}
& \other{58.2}
& \other{66.6}
& \other{74.2}
& \other{58.7}
& \other{86.8}
& \other{69.4}
& \other{69.0}
& \other{56.0}
\\

\cmidrule{2-25}

& SPM\cite{wu2025efficient}
& \other{18.3}
& \other{82.8}
& \other{70.7}
& \other{87.3}
& \other{95.5}
& \other{72.4}
& \other{81.4}
& \other{78.8}
& \other{75.9}
& \other{72.4}
& \other{58.4}
& \other{58.9}
& \other{65.9}
& \other{56.5}
& \other{63.7}
& \other{60.2}
& \other{69.2}
& \other{73.2}
& \second{65.4}
& \other{83.1}
& \other{69.4}
& \other{72.1}
& \other{54.3}
\\

& \cellcolor{spikeblue}\textbf{w/ SpikePEFT (Ours)}
& \cellcolor{spikeblue}\best{6.0}
& \cellcolor{spikeblue}\second{83.8}\ \textcolor{gain}{(+\ 1.0)}
& \cellcolor{spikeblue}\other{71.3}\ \textcolor{gain}{(+\ 0.6)}
& \cellcolor{spikeblue}\other{86.8}
& \cellcolor{spikeblue}\other{93.8}
& \cellcolor{spikeblue}\second{72.6}
& \cellcolor{spikeblue}\second{81.6}
& \cellcolor{spikeblue}\best{86.2}
& \cellcolor{spikeblue}\other{71.4}
& \cellcolor{spikeblue}\best{73.8}
& \cellcolor{spikeblue}\second{58.6}
& \cellcolor{spikeblue}\best{66.8}
& \cellcolor{spikeblue}\other{60.2}
& \cellcolor{spikeblue}\second{57.4}
& \cellcolor{spikeblue}\best{64.1}
& \cellcolor{spikeblue}\best{60.6}
& \cellcolor{spikeblue}\second{70.1}
& \cellcolor{spikeblue}\best{77.4}
& \cellcolor{spikeblue}\other{63.5}
& \cellcolor{spikeblue}\best{89.9}
& \cellcolor{spikeblue}\other{64.0}
& \cellcolor{spikeblue}\other{71.0}
& \cellcolor{spikeblue}\second{56.2}
\\

& E-3DSNN-L + SVL\cite{qiu2025svl}
& \other{20.1}
& \other{83.3}
& \second{71.8}
& \second{91.7}
& \second{98.5}
& \other{69.0}
& \other{80.3}
& \other{81.8}
& \other{75.3}
& \other{73.4}
& \other{58.4}
& \second{66.6}
& \best{66.3}
& \other{55.8}
& \second{63.9}
& \other{59.4}
& \other{67.9}
& \second{77.2}
& \other{63.7}
& \other{85.5}
& \second{69.6}
& \best{73.3}
& \best{58.4}
\\

& \cellcolor{spikeblue}\textbf{w/ SpikePEFT (Ours)}
& \cellcolor{spikeblue}\second{6.1}
& \cellcolor{spikeblue}\best{83.9}\ \textcolor{gain}{(+\ 0.6)}
& \cellcolor{spikeblue}\best{72.4}\ \textcolor{gain}{(+\ 0.6)}
& \cellcolor{spikeblue}\best{91.9}
& \cellcolor{spikeblue}\best{98.7}
& \cellcolor{spikeblue}\best{73.4}
& \cellcolor{spikeblue}\best{82.2}
& \cellcolor{spikeblue}\second{83.9}
& \cellcolor{spikeblue}\best{77.6}
& \cellcolor{spikeblue}\second{73.6}
& \cellcolor{spikeblue}\best{62.8}
& \cellcolor{spikeblue}\other{61.8}
& \cellcolor{spikeblue}\other{63.6}
& \cellcolor{spikeblue}\best{61.9}
& \cellcolor{spikeblue}\other{62.4}
& \cellcolor{spikeblue}\other{58.8}
& \cellcolor{spikeblue}\best{73.4}
& \cellcolor{spikeblue}\other{76.9}
& \cellcolor{spikeblue}\best{65.6}
& \cellcolor{spikeblue}\second{87.4}
& \cellcolor{spikeblue}\best{69.8}
& \cellcolor{spikeblue}\other{69.5}
& \cellcolor{spikeblue}\other{52.8}
\\

\noalign{\vskip -0.5ex}
\bottomrule

\end{tabular}%
}
\vspace{-6pt}
\end{table*}
\begin{table*}[!t]
\centering

\caption{
Semantic segmentation results on S3DIS 6-fold\cite{Armeni_2016_CVPR},
ScanNet200 Val\cite{rozenberszki2022language},
ScanNet++ Val\cite{yeshwanth2023scannet++},
and NuScenes Val\cite{caesar2020nuscenes}. \\
Trainable parameters are consistent with Table~\ref{tab:scannet}.
}
\label{tab:large_seg}

\fontsize{5.0}{6.0}\selectfont
\setlength{\tabcolsep}{1pt}
\renewcommand{\arraystretch}{0.95}

\resizebox{\textwidth}{!}{%
\begin{tabular}{cl|ccc|ccc|ccc|ccc}
\toprule

\multirow[c]{2}{*}{Type}
&
\multirow[c]{2}{*}{Method}
&
\multicolumn{3}{c|}{\raisebox{0.35ex}{S3DIS 6-fold\cite{Armeni_2016_CVPR}}}
&
\multicolumn{3}{c|}{\raisebox{0.35ex}{ScanNet200 Val\cite{rozenberszki2022language}}}
&
\multicolumn{3}{c|}{\raisebox{0.35ex}{ScanNet++ Val\cite{yeshwanth2023scannet++}}}
&
\multicolumn{3}{c}{\raisebox{0.35ex}{NuScenes Val\cite{caesar2020nuscenes}}}
\\

\cline{3-5}
\cline{6-8}
\cline{9-11}
\cline{12-14}

&
&
\raisebox{-0.45ex}{mAcc}
&
\raisebox{-0.45ex}{allAcc}
&
\raisebox{-0.45ex}{mIoU}
&
\raisebox{-0.45ex}{mAcc}
&
\raisebox{-0.45ex}{allAcc}
&
\raisebox{-0.45ex}{mIoU}
&
\raisebox{-0.45ex}{mAcc}
&
\raisebox{-0.45ex}{allAcc}
&
\raisebox{-0.45ex}{mIoU}
&
\raisebox{-0.45ex}{mAcc}
&
\raisebox{-0.45ex}{allAcc}
&
\raisebox{-0.45ex}{mIoU}
\\
\midrule


\multirow[c]{5}{*}{ANN}

& Point-MAE\cite{pang2022pointmae}
& 82.3 & 90.5 & 75.1
& 44.7 & 82.5 & 34.3
& 53.7 & 81.8 & 39.9
& 85.3 & 94.3 & 78.7
\\

& PointNeXt\cite{qian2022pointnext}
& 81.5 & 89.8 & 74.4
& 44.1 & 82.2 & 33.7
& 53.2 & 81.5 & 39.4
& 84.9 & 94.1 & 78.3
\\

& ReCon\cite{qi2023recon}
& 82.6 & 90.6 & 75.7
& 45.8 & 82.8 & 35.2
& 54.6 & 82.0 & 40.6
& 85.8 & 94.5 & 79.3
\\

& PointMamba\cite{liang2024pointmamba}
& 83.3 & 90.9 & 76.3
& 46.5 & 83.1 & 35.8
& 53.5 & 82.3 & 41.1
& 86.7 & 94.7 & 79.9
\\

& PointATA\cite{sun2026align}
& 83.0 & 90.7 & 75.9
& 46.0 & 82.9 & 35.4
& 54.9 & 82.1 & 40.8
& 86.1 & 94.5 & 79.5
\\


\midrule

\multirow[c]{7}{*}{SNN}

& P2SResLNet-B\cite{wu2024point}
& \other{79.5}
& \other{89.7}
& \other{72.1}
& \other{39.8}
& \other{80.9}
& \other{29.7}
& \other{50.4}
& \other{81.7}
& \other{37.0}
& \other{80.6}
& \other{92.8}
& \other{73.5}
\\

& E-3DSNN-L\cite{qiu2025efficient}
& \other{80.8}
& \other{90.1}
& \other{73.4}
& \other{40.7}
& \other{81.4}
& \other{30.8}
& \other{51.8}
& \other{82.1}
& \other{37.8}
& \other{81.8}
& \other{93.3}
& \other{74.9}
\\

& SPT\cite{wu2025spiking}
& \other{82.6}
& \best{91.9}
& \other{75.2}
& \best{45.4}
& \other{81.9}
& \other{32.1}
& \other{53.2}
& \best{84.0}
& \second{39.6}
& \best{86.0}
& \other{93.8}
& \other{76.2}
\\

\cmidrule{2-14}

& SPM\cite{wu2025efficient}
& \second{84.6}
& \other{91.5}
& \other{77.1}
& \other{44.3}
& \other{82.7}
& \second{34.1}
& \other{55.8}
& \other{83.6}
& \other{39.1}
& \second{85.2}
& \second{94.6}
& \best{78.3}
\\

& \cellcolor{spikeblue}\textbf{w/ SpikePEFT (Ours)}
& \cellcolor{spikeblue}\best{85.0}\textcolor{gain}{(+\ 0.4)}
& \cellcolor{spikeblue}\second{91.8}\textcolor{gain}{(+\ 0.3)}
& \cellcolor{spikeblue}\best{77.7}\textcolor{gain}{(+\ 0.6)}
& \cellcolor{spikeblue}\second{45.1}\textcolor{gain}{(+\ 0.8)}
& \cellcolor{spikeblue}\best{83.0}\textcolor{gain}{(+\ 0.3)}
& \cellcolor{spikeblue}\best{34.8}\textcolor{gain}{(+\ 0.7)}
& \cellcolor{spikeblue}\best{56.6}\textcolor{gain}{(+\ 0.8)}
& \cellcolor{spikeblue}\second{83.9}\textcolor{gain}{(+\ 0.3)}
& \cellcolor{spikeblue}\best{39.8}\textcolor{gain}{(+\ 0.7)}
& \cellcolor{spikeblue}\other{85.0}\textcolor{drop}{(-\ 0.2)}
& \cellcolor{spikeblue}\best{94.7}\textcolor{gain}{(+\ 0.1)}
& \cellcolor{spikeblue}\second{78.1}\textcolor{drop}{(-\ 0.2)}
\\

\cmidrule{2-14}

& E-3DSNN-L + SVL\cite{qiu2025svl}
& \other{82.1}
& \other{90.7}
& \other{76.9}
& \other{43.6}
& \other{82.5}
& \other{33.5}
& \other{55.5}
& \other{83.7}
& \other{38.8}
& \other{84.4}
& \other{94.3}
& \other{77.6}
\\

& \cellcolor{spikeblue}\textbf{w/ SpikePEFT (Ours)}
& \cellcolor{spikeblue}\other{82.4}\textcolor{gain}{(+\ 0.3)}
& \cellcolor{spikeblue}\other{91.1}\textcolor{gain}{(+\ 0.4)}
& \cellcolor{spikeblue}\second{77.3}\textcolor{gain}{(+\ 0.4)}
& \cellcolor{spikeblue}\other{44.3}\textcolor{gain}{(+\ 0.7)}
& \cellcolor{spikeblue}\second{82.9}\textcolor{gain}{(+\ 0.4)}
& \cellcolor{spikeblue}\other{33.9}\textcolor{gain}{(+\ 0.4)}
& \cellcolor{spikeblue}\second{56.0}\textcolor{gain}{(+\ 0.5)}
& \cellcolor{spikeblue}\best{84.0}\textcolor{gain}{(+\ 0.3)}
& \cellcolor{spikeblue}\other{39.3}\textcolor{gain}{(+\ 0.5)}
& \cellcolor{spikeblue}\other{84.4}\textcolor{gain}{(+\ 0.0)}
& \cellcolor{spikeblue}\second{94.6}\textcolor{gain}{(+\ 0.3)}
& \cellcolor{spikeblue}\other{78.0}\textcolor{gain}{(+\ 0.4)}
\\

\noalign{\vskip -0.5ex}
\bottomrule

\end{tabular}
}

\end{table*}

\begin{figure}[!t]
\centering
\includegraphics[width=1\columnwidth]{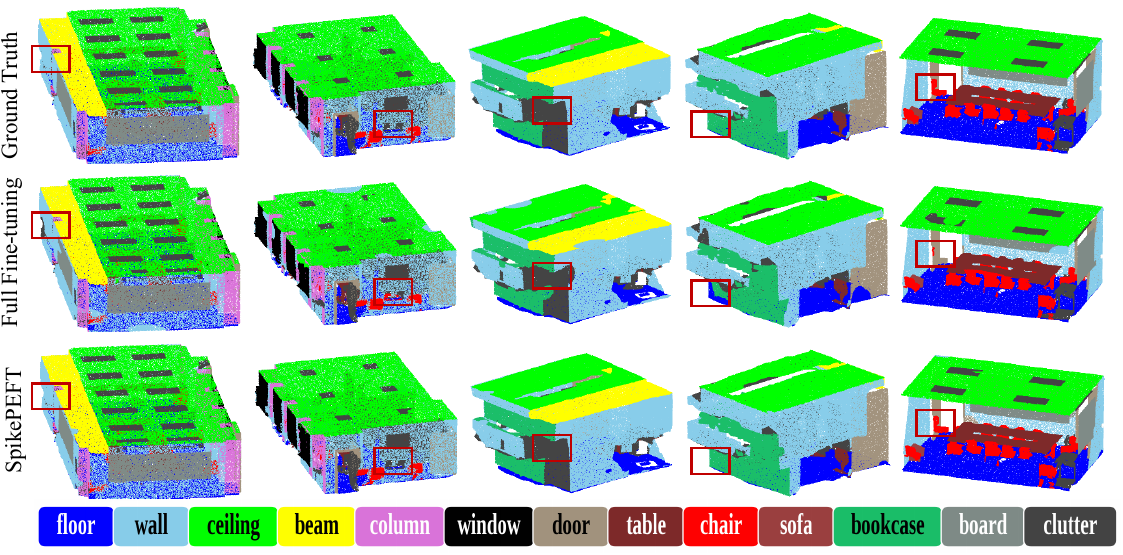}
\caption{Qualitative analysis results for semantic segmentation on S3DIS.}
\label{fig:scene}
\vspace{-16pt}
\end{figure}

\begin{figure*}[!t]
\centering
\includegraphics[width=1\textwidth]{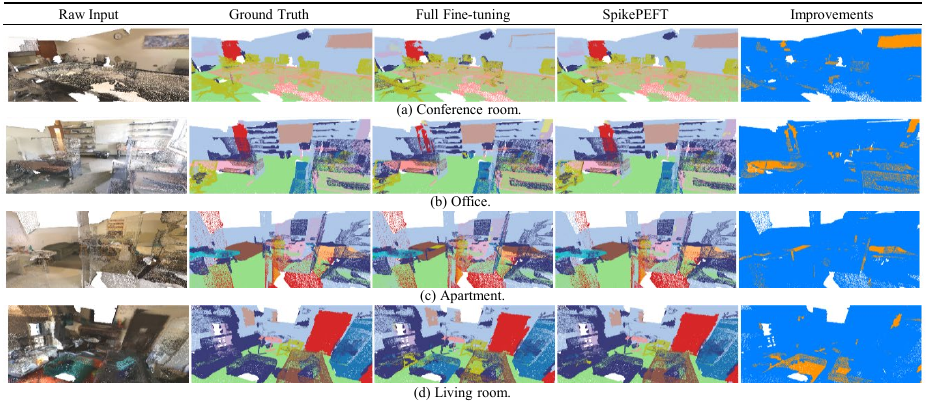}
\vspace{-12pt}
\caption{Qualitative comparison of semantic segmentation 
results on ScanNet. 
Prediction regions improved by SpikePEFT are highlighted for clarity.}
\label{fig:scannet}
\vspace{-3pt}
\end{figure*}

\noindent\textbf{Indoor Semantic Segmentation.}
The \textit{S3DIS}\cite{Armeni_2016_CVPR} is a 
widely used indoor scene semantic segmentation benchmark 
comprising densely annotated room-scale point clouds, 
evaluated under both the Area-5 split and 6-fold 
cross-validation protocols. 
The \textit{ScanNet V2}\cite{dai2017scannet} 
is a large-scale indoor 3D scene benchmark 
with dense point-wise annotations across 20 semantic categories. 
As shown in Tables~\ref{tab:semantic_segmentation},
\ref{tab:scannet}, and~\ref{tab:large_seg},
SpikePEFT consistently improves dense indoor prediction.
On S3DIS Area 5, it improves SPM from 63.3\% to 64.0\% mIoU,
while on S3DIS 6-fold the mIoU further increases from 77.1\% to 77.7\%.
On ScanNet V2, the mIoU of SPM and E-3DSNN-L with SVL consistently increases
from 70.7\% and 71.8\% to 71.3\% and 72.4\%, respectively.
The visualization results in Figures~\ref{fig:scannet} and~\ref{fig:scene}
further show that SpikePEFT yields more complete and spatially consistent
semantic predictions.

\noindent\textbf{Large-Scale Semantic Segmentation.}
The \textit{ScanNet200}\cite{rozenberszki2022language} 
extends ScanNet to 200 semantic categories, providing 
a substantially more fine-grained and long-tailed 
benchmark for large-scale 3D scene 
understanding. The \textit{ScanNet++}\cite{yeshwanth2023scannet++} 
features high-fidelity indoor reconstructions with rich 
geometric details and fine-grained semantic annotations, 
posing additional challenges for dense 3D prediction. 
As shown in Table~\ref{tab:large_seg}, SpikePEFT delivers notable
performance on both ScanNet200 and ScanNet++, demonstrating that
parameter-efficient adaptation remains effective under increased semantic
granularity and scene complexity. This is particularly noteworthy for PEFT,
where large-scale dense 3D prediction typically demands substantial adaptation capacity.

\begin{figure}[!t]
\centering
\includegraphics[width=1\columnwidth]{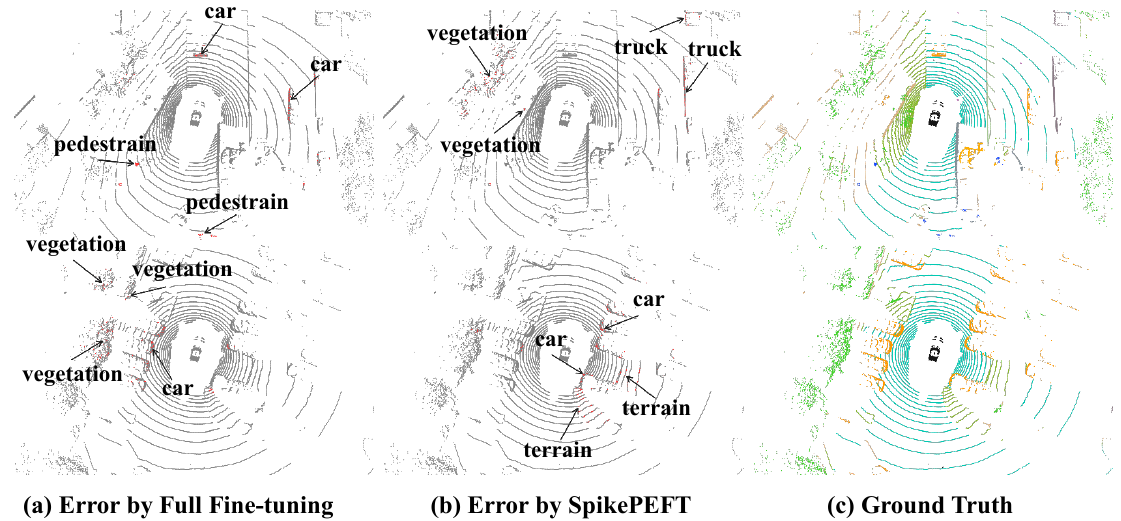}
\caption{Qualitative analysis results for semantic segmentation on NuScenes.}
\label{fig:nuscenes}
\end{figure}

\noindent\textbf{Outdoor Semantic Segmentation.}
The \textit{NuScenes}\cite{caesar2020nuscenes} 
is a large-scale autonomous driving benchmark 
featuring outdoor LiDAR point clouds, widely 
used for scene-level semantic segmentation and 
3D object detection in complex urban traffic environments. 
As noted in prior studies\cite{wu2025sonata,Liu_2026_CVPR}, indoor and outdoor point-cloud 
datasets exhibit a pronounced domain gap due to substantial 
differences in point density, spatial scale, and scene structure. 
As shown in Table~\ref{tab:large_seg}, SpikePEFT maintains 
competitive performance under this 
substantially different outdoor data distribution. 
Figure~\ref{fig:nuscenes} further compares the prediction errors of
full fine-tuning and SpikePEFT against the ground truth.
SpikePEFT exhibits a similar error pattern to full fine-tuning across
different semantic regions, indicating a comparable capability for outdoor scene understanding.

\subsection{Few-shot Learning} 

We further evaluate SpikePEFT under the few-shot setting on a dedicated benchmark derived from ModelNet40\cite{wu20153d}. Following previous
point cloud foundation model 
studies\cite{pang2022pointmae,wang2026pointrft,
chen2023pointgpt}, we employ the n-way, m-shot configuration 
with $n \in \{5,10\}$ and $m \in \{10,20\}$. 
For each setting, we report the average accuracy (\%) ± standard deviation (\%) of 
10 independent experiments without voting. 
Table~\ref{tab:fewshot} shows that SpikePEFT achieves the best
or second-best accuracy in most cases. 
This consistent trend demonstrates the generalization capability of SpikePEFT under limited training data regimes.

\begin{table}[!ht]
\vspace{-12pt}
\caption{
Few-shot learning on ModelNet40\cite{wu20153d}.}
\vspace{-4pt}
\centering
\scriptsize
\setlength{\tabcolsep}{4pt}
\renewcommand{\arraystretch}{1.1}

\begin{tabular}{clcccc}
\toprule
\multirow{2}{*}{Type}
&
\multirow{2}{*}{Method}
&
\multicolumn{2}{c}{5-way}
&
\multicolumn{2}{c}{10-way}
\\
\cmidrule(lr){3-4}
\cmidrule(lr){5-6}
&
&
10-shot
&
20-shot
&
10-shot
&
20-shot
\\
\midrule

\multirow[c]{5}{*}{ANN}
& Point-BERT\cite{yu2022point}
& 94.6$\pm$3.1
& 96.3$\pm$2.7
& 91.0$\pm$5.4
& 92.7$\pm$5.1
\\

& MaskPoint\cite{liu2022masked}
& 95.0$\pm$3.7
& 97.2$\pm$1.7
& 91.4$\pm$4.0
& 93.4$\pm$3.5
\\

& ReCon\cite{qi2023recon}
& 97.3$\pm$1.9
& 98.9$\pm$1.2
& 93.3$\pm$3.9
& 93.3$\pm$3.9
\\

& PointGPT-S\cite{chen2023pointgpt}
& 98.0$\pm$1.9
& 99.0$\pm$1.0
& 94.1$\pm$3.3
& 96.1$\pm$2.8
\\

& ACT\cite{dongautoencoders}
& 96.8$\pm$2.3
& 98.0$\pm$1.4
& 93.3$\pm$4.0
& 95.6$\pm$2.8
\\

\midrule

\multirow[c]{8}{*}{SNN}
& Spike PointNet\cite{lan2023efficient}
& 91.8$\pm$4.5
& 96.2$\pm$2.3
& 87.5$\pm$5.8
& 91.6$\pm$4.4
\\

& E-3DSNN-L\cite{qiu2025efficient}
& 94.2$\pm$3.1
& 97.4$\pm$1.8
& 90.5$\pm$4.7
& 93.8$\pm$3.8
\\

& SPT\cite{wu2025spiking}
& 94.8$\pm$2.9
& 97.8$\pm$1.6
& 91.2$\pm$4.5
& 94.3$\pm$3.5
\\

\cmidrule(l){2-6}

& SPM\cite{wu2025efficient}
& 94.5$\pm$2.8
& 97.2$\pm$1.9
& 90.8$\pm$4.8
& 93.6$\pm$3.9
\\

& \cellcolor{spikeblue}\textbf{w/ SpikePEFT (Ours)}
& \cellcolor{spikeblue}94.7$\pm$2.1
& \cellcolor{spikeblue}\underline{98.4$\pm$1.4}
& \cellcolor{spikeblue}\underline{92.7$\pm$4.1}
& \cellcolor{spikeblue}\textbf{95.3$\pm$3.2}
\\[1.5pt]

\cmidrule(l){2-6}

& E-3DSNN-L + SVL\cite{qiu2025svl}
& \underline{95.1$\pm$2.6}
& 97.8$\pm$1.7
& 91.7$\pm$4.5
& 94.2$\pm$3.6
\\

& \cellcolor{spikeblue}\textbf{w/ SpikePEFT (Ours)}
& \cellcolor{spikeblue}\textbf{96.7$\pm$2.0}
& \cellcolor{spikeblue}\textbf{98.6$\pm$1.3}
& \cellcolor{spikeblue}\textbf{93.1$\pm$3.8}
& \cellcolor{spikeblue}\underline{95.1$\pm$3.0}
\\[1.5pt]

\bottomrule
\end{tabular}

\label{tab:fewshot}
\vspace{-8pt}
\end{table}
\begin{table}[t]
\caption{Ablation study on the settings of SpikePEFT.}
\vspace{-4pt}
\centering
\setlength{\tabcolsep}{5pt}
\renewcommand{\arraystretch}{1}
\small

\resizebox{\linewidth}{!}{
\begin{tabular}{c ccc c cccc}
\toprule
\multirow{2}{*}{Setting}
&
\multicolumn{2}{c}{IDT}
&
\multirow{2}{*}{SSDA}
&
\multirow{2}{*}{\#TP (M)}
&
\multirow{2}{*}{MN40}
&
\multirow{2}{*}{BG}
&
\multirow{2}{*}{ONLY}
&
\multirow{2}{*}{RS}
\\
\cmidrule(lr){2-3}
&
$\Delta_{\lambda}$
& $\Delta_{v}$
&
&
&
&
&
&
\\
\midrule

--
& \multicolumn{3}{c}{Full fine-tuning}
& 12.81
& \underline{92.3}
& 90.2
& 89.5
& 84.2
\\

--
& \multicolumn{3}{c}{Linear probing}
& 0.35
& 86.7
& 79.8
& 78.9
& 72.6
\\
\midrule

A0
& \checkmark
&
&
&
0.42
& 87.9
& 81.5
& 80.7
& 74.9
\\

A1
&
& \checkmark
&
&
0.42
& 87.6
& 81.1
& 80.3
& 74.5
\\

A2
& \checkmark
& \checkmark
&
&
0.49
& 89.1
& 83.6
& 82.8
& 77.8
\\

A3
&
&
& \checkmark
&
0.53
& 91.7
& 89.7
& 89.2
& 83.9
\\

A4
& \checkmark
&
& \checkmark
&
0.60
& 92.1
& \underline{90.6}
& 89.8
& 84.8
\\

A5
&
& \checkmark
& \checkmark
&
0.60
& 92.0
& 90.4
& \underline{90.1}
& \underline{85.0}
\\

\cellcolor{spikeblue}\textbf{A6}
& \cellcolor{spikeblue}\checkmark
& \cellcolor{spikeblue}\checkmark
& \cellcolor{spikeblue}\checkmark
& \cellcolor{spikeblue}0.67
& \cellcolor{spikeblue}\textbf{92.4}
& \cellcolor{spikeblue}\textbf{91.1}
& \cellcolor{spikeblue}\textbf{90.8}
& \cellcolor{spikeblue}\textbf{85.6}
\\

\bottomrule
\end{tabular}
}
\label{tab:ablation_components}
\end{table}
\subsection{Ablation Study}

Unless otherwise stated, all experiments are based on SPM to investigate the rationale and effectiveness of SpikePEFT. 

\noindent\textbf{Ablation on different components.}
We first study the contribution of each component in SpikePEFT.
As illustrated in Table~\ref{tab:ablation_components},
SSDA constitutes the primary source of performance
improvement (A3), whereas IDT alone yields moderate gains (A0--A2).
Building on SSDA, fine-tuning membrane
decay or firing threshold brings further gains (A4--A5), and their
combination achieves the best result (A6).
Overall, SSDA exploits task-relevant information
hidden in silent states, while IDT complements it by
adjusting neuron-intrinsic dynamics.

\begin{figure*}[!t]
\centering
\includegraphics[width=1\textwidth]{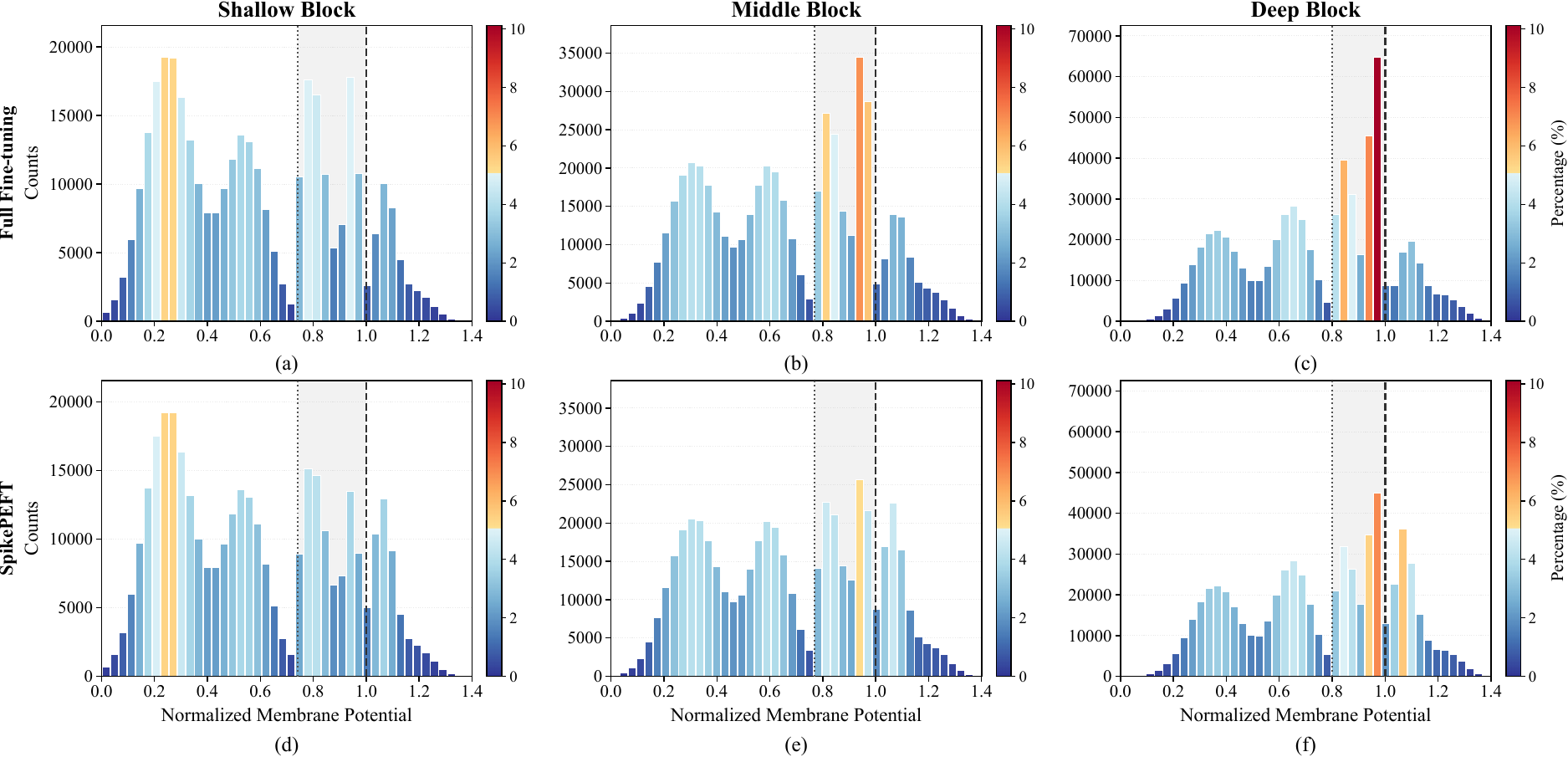}
\caption{Normalized membrane potential distributions in the shallow, middle, and deep blocks of SPM under full fine-tuning and SpikePEFT on ModelNet40 with $T=4$. For clearer visualization, the firing threshold is normalized to 1, and the shaded region denotes the sub-threshold interval considered by SSDA.
\label{fig:distribution}
}
\end{figure*}

\begin{figure}[!t]
\centering
\includegraphics[width=1\columnwidth]{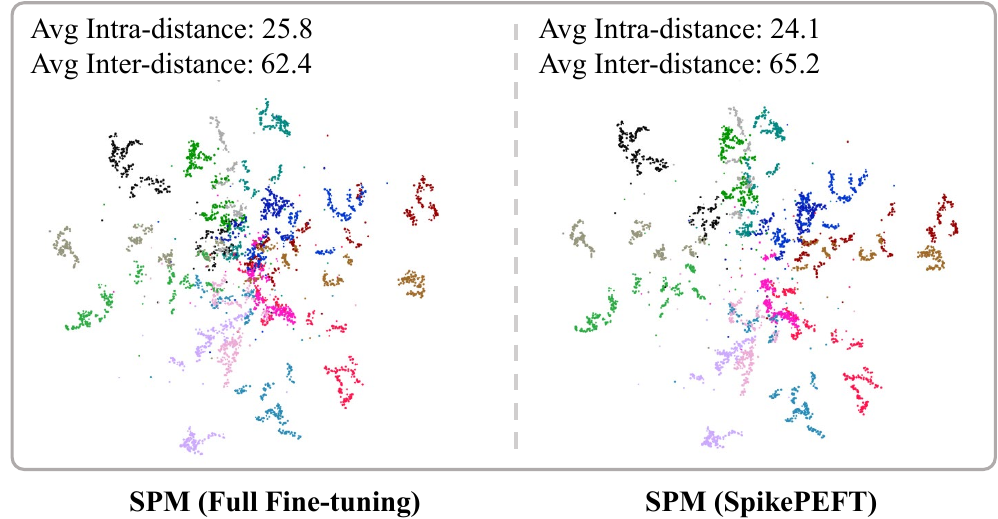}
\caption{Comparison of t-SNE feature clustering between full fine-tuning and SpikePEFT, with average intra- and inter-class distances reported.}
\label{fig:tsne}
\end{figure}

\begin{table}[t]
\vspace{-8pt}
\caption{Ablation study of time step. ANN refers to PointMamba, SNN refers to SPM.}

\centering
\setlength{\tabcolsep}{6pt}
\renewcommand{\arraystretch}{1.05}
\small
\begin{tabular}{ccccccc}
\toprule
Type
& $T$
& Energy (mJ)
& MN40
& BG
& ONLY
& RS
\\
\midrule

ANN
& -- & 18.9 & 92.4 & 90.2 & 89.6 & 85.4
\\

\midrule

\multirow{5}{*}{SNN}
& 1 & 1.5 & 91.6 & 88.9 & 87.8 & 83.3
\\
& 2 & 2.8 & 91.8 & 89.8 & 88.6 & 83.7
\\
& 3 & 3.9 & 92.1 & 90.2 & 89.2 & 83.8
\\
& 4 & 5.4 & 92.3 & 90.2 & 89.5 & 84.2
\\
& 6 & 7.6 & 92.3 & 90.0 & 89.6 & 84.3
\\

\midrule

\multirow{5}{*}{PEFT}
& 1 & 1.8 & 91.9 & 89.9 & 89.3 & 84.1
\\
& 2 & 3.3 & 92.1 & 90.5 & 89.9 & 84.3
\\
& 3 & 4.6 & 92.3 & \underline{90.9}  & 90.5 & 85.4
\\
& \cellcolor{spikeblue}4 & \cellcolor{spikeblue}6.2 
& \cellcolor{spikeblue}\textbf{92.4} 
& \cellcolor{spikeblue}\textbf{91.1} 
& \cellcolor{spikeblue}\textbf{90.8} 
& \cellcolor{spikeblue}\underline{85.6}
\\
& 6 & 8.7 
& \underline{92.3} 
& \underline{90.9} 
& \underline{90.6} 
& \textbf{85.7}
\\

\bottomrule
\end{tabular}
\label{tab:timestep_energy}
\vspace{-8pt}
\end{table}

\noindent\textbf{Ablation on time steps.}
In SNNs, the number of time steps affects temporal information
accumulation. We conduct experiments 
with different numbers of time steps and report the corresponding 
accuracy and energy consumption 
in Table~\ref{tab:timestep_energy}. 
SpikePEFT consistently improves performance within 
relatively short temporal windows. By releasing 
information retained in silent states, SSDA compensates 
for insufficient spike responses and reduces the reliance 
on long temporal windows. 

\noindent\textbf{Analysis of membrane potential distributions.}
Figure~\ref{fig:distribution} compares the membrane potential
distributions of SPM under full fine-tuning and SpikePEFT.
The shallow block remains largely unchanged, whereas clearer
redistribution emerges in the middle and deep blocks, particularly
around the firing threshold. This layer-dependent behavior indicates
that SpikePEFT does not indiscriminately amplify neuronal activity. 
Instead, it selectively reshapes task-relevant near-threshold states
in higher-level representations. This recovers task-relevant
information suppressed by binary spike generation,
providing richer evidence for downstream adaptation.

\begin{figure}[!t]
\centering
\includegraphics[width=1\columnwidth]{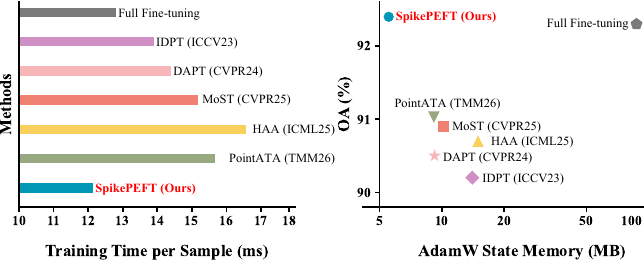} 
\caption{Comparison of training efficiency across different PEFT methods.}
\label{fig:efficiency}
\vspace{-8pt}
\end{figure}

\noindent\textbf{Analysis of computational efficiency.}
In SNNs, spike-driven computation is dominated by accumulation (AC)
operations, whereas dense continuous computation relies on
multiply--accumulate (MAC) operations. FR denotes the average firing rate.
According to the research\cite{6757323}, a 32-bit floating-point 
consumes 4.6\,pJ for a MAC operation and 0.9\,pJ for an AC 
operation, namely 
$E_{\text{MAC}} = 4.6\,\text{pJ}$ and $E_{\text{AC}} = 0.9\,\text{pJ}$.
As shown in Table~\ref{tab:operation_energy}, ANN-based PEFT 
methods introduce substantial additional MAC operations, 
whereas SpikePEFT mainly incurs low-cost AC operations. 
IDT introduces 
negligible overhead, while SSDA accounts for most of the 
additional AC operations. SpikePEFT consumes 6.22~mJ, 14.8\% higher than SPM but 67.1\% lower than PointMamba, 
with only a slight increase in firing rate. Consequently, 
the performance gains of SpikePEFT are achieved without 
compromising the inherent spike sparsity and energy efficiency of SNNs. 
By combining lightweight trainable parameters with spike-driven adaptation,
SpikePEFT reduces optimization overhead while maintaining
competitive accuracy. In particular, it reduces AdamW state memory by
94.7\% compared with full fine-tuning and achieves the shortest training
time among the compared methods, as illustrated in
Figure~\ref{fig:efficiency}. Overall, SpikePEFT achieves the best balance
between accuracy and computational efficiency.
\begin{table}[t]
\caption{Analysis of neuronal activity and computational efficiency
on ModelNet40 with $T=4$.}
\centering
\small
\setlength{\tabcolsep}{3pt}
\renewcommand{\arraystretch}{1}
\begin{tabular}{lccccc}
\toprule
Method
& \makecell{AC (G)}
& \makecell{MAC (G)}
& \makecell{FR}
& \makecell{Energy}
& \makecell{OA (\%)} \\
\midrule

PointMamba
& 0
& 4.112
& --
& 18.92
& 92.4
\\

\midrule
\multicolumn{6}{c}{\textit{ANN-PEFT Protocol}} \\
\midrule

SPM
& 5.78
& 0.047
& 0.178
& 5.42
& 92.3
\\

w/ IDPT\cite{zha2023idpt}
& 6.02
& 1.070
& 0.175
& 10.34
& 90.2
\\

w/ DAPT\cite{zhou2024dapt}
& 6.08
& 1.232
& 0.182
& 11.14
& 90.5
\\

w/ MoST\cite{han2025most}
& 6.16
& 1.523
& 0.177
& 12.55
& 90.9
\\

w/ HAA\cite{haa2025}
& 6.25
& 2.170
& 0.180
& 15.61
& 90.7
\\

w/ PointATA\cite{sun2026align}
& 6.11
& 1.427
& 0.189
& 12.06
& 91.0
\\

\midrule
\multicolumn{6}{c}{\textit{SpikePEFT Protocol}} \\
\midrule

SPM
& 5.78
& 0.047
& 0.178
& 5.42
& 92.3
\\

w/ IDT
& 5.86
& 0.047
& 0.175
& 5.49
& 89.1
\\

w/ SSDA
& 6.46
& 0.047
& 0.182
& 6.03
& 91.7
\\

\cellcolor{spikeblue}\textbf{w/ SpikePEFT}
& \cellcolor{spikeblue}6.67
& \cellcolor{spikeblue}0.047
& \cellcolor{spikeblue}0.186
& \cellcolor{spikeblue}6.22
& \cellcolor{spikeblue}\textbf{92.4}
\\

\bottomrule
\end{tabular}
\label{tab:operation_energy}
\end{table}
\section{Conclusion}

In this paper, we propose SpikePEFT, 
the first parameter-efficient fine-tuning 
framework for spiking point cloud models. 
We reveal the limitations of existing downstream 
adaptation strategies for spiking point cloud models, 
and address them through neuron-intrinsic dynamics 
adaptation and silent-state disambiguation. 
Extensive experiments across diverse benchmarks 
demonstrate that SpikePEFT 
achieves competitive performance compared with full fine-tuning while 
significantly reducing the number of trainable parameters. 
As an early exploration of PEFT for spiking foundation 
models with generic neuron-level mechanisms, we hope SpikePEFT can serve as a strong 
baseline and inspire future research on efficient adaptation 
across broader neuromorphic vision architectures.


\bibliographystyle{IEEEtran}
\bibliography{references}

@inproceedings{qi2017pointnet,
  title     = {PointNet: Deep Learning on Point Sets for 3D Classification and Segmentation},
  author    = {Qi, Charles R. and Su, Hao and Mo, Kaichun and Guibas, Leonidas J.},
  booktitle = {CVPR},
  year      = {2017}
}

@inproceedings{qian2022pointnext,
  title     = {PointNeXt: Revisiting PointNet++ with Improved Training and Scaling Strategies},
  author    = {Qian, Guocheng and Li, Xiuwei and Peng, Houwen and Mai, Jiahui and Hammoud, Hassan and Elhoseiny, Mohamed and Ghanem, Bernard},
  booktitle = {NeurIPS},
  year      = {2022}
}

@inproceedings{yu2022point,
  title={Point-bert: Pre-training 3d point cloud transformers with masked point modeling},
  author={Yu, Xumin and Tang, Lulu and Rao, Yongming and Huang, Tiejun and Zhou, Jie and Lu, Jiwen},
  booktitle={CVPR},
  year={2022}
}

@inproceedings{pang2022pointmae,
  title     = {Masked Autoencoders for Point Cloud Self-Supervised Learning},
  author    = {Pang, Yatian and Wang, Wenxiao and Tay, Francis E. H. and Liu, Wei and Tian, Yijun and Yuan, Li},
  booktitle = {ECCV},
  year      = {2022}
}

@article{chen2023pointgpt,
  title={Pointgpt: Auto-regressively generative pre-training from point clouds},
  author={Chen, Guangyan and Wang, Meiling and Yang, Yi and Yu, Kai and Yuan, Li and Yue, Yufeng},
  journal={NeurIPS},
  year={2023}
}

@inproceedings{gu2023mamba,
  title={Mamba: Linear-time sequence modeling with selective state spaces},
  author={Gu, Albert and Dao, Tri},
  booktitle={COLM},
  year = {2024}
}

@inproceedings{liang2024pointmamba,
  title     = {PointMamba: A Simple State Space Model for Point Cloud Analysis},
  author    = {Liang, Dingkang and Zhou, Xin and Xu, Wei and Zhu, Xingkui and Zou, Zhikang and Ye, Xiaoqing and Tan, Xiao and Bai, Xiang},
  booktitle = {NeurIPS},
  year      = {2024}
}

@inproceedings{zha2023idpt,
  title     = {Instance-aware Dynamic Prompt Tuning for Pre-trained Point Cloud Models},
  author    = {Zha, Yaohua and Wang, Jinpeng and Dai, Tao and Chen, Bin and Wang, Zhi and Xia, Shu-Tao},
  booktitle = {ICCV},
  year      = {2023}
}

@inproceedings{zhou2024dapt,
  title     = {Dynamic Adapter Meets Prompt Tuning: Parameter-Efficient Transfer Learning for Point Cloud Analysis},
  author    = {Zhou, Xin and Liang, Dingkang and Xu, Wei and Zhu, Xingkui and Xu, Yihan and Zou, Zhikang and Bai, Xiang},
  booktitle = {CVPR},
  year      = {2024}
}

@inproceedings{haa2025,
  title     = {Exploring Vision Semantic Prompt for Efficient Point Cloud Understanding},
  author    = {Zha, Yixin and Wang, Chuxin and Yang, Wenfei and Zhang, Tianzhu and Wu, Feng},
  booktitle = {ICML},
  year      = {2025}
}

@inproceedings{wang2025pointlora,
  title     = {PointLoRA: Low-Rank Adaptation with Token Selection for Point Cloud Learning},
  author    = {Wang, Song and Liu, Xiaolu and Kong, Lingdong and Xu, Jianyun and Hu, Chunyong and Fang, Gongfan and Li, Wentong and Zhu, Jianke and Wang, Xinchao},
  booktitle = {CVPR},
  year      = {2025}
}

@inproceedings{qi2023recon,
    title     = {Contrast with Reconstruct: Contrastive 3D Representation Learning Guided by Generative Pretraining},
    author    = {Qi, Zekun and Dong, Runpei and Fan, Guofan and Ge, Zheng and Zhang, Xiangyu and Ma, Kaisheng and Yi, Li},
    booktitle = {ICML},
    year      = {2023}
}

@article{wang2019dgcnn,
    title={Dynamic Graph CNN for Learning on Point Clouds},
    author={Wang, Yue and Sun, Yongbin and Liu, Ziwei and Sarma, Sanjay E. and Bronstein, Michael M. and Solomon, Justin M.},
    journal={TOG},
    year={2019},
    publisher={ACM}
}

@inproceedings{houlsby2019parameter,
  title={Parameter-efficient transfer learning for NLP},
  author={Houlsby, Neil and Giurgiu, Andrei and Jastrzebski, Stanislaw and Morrone, Bruna and De Laroussilhe, Quentin and Gesmundo, Andrea and Attariyan, Mona and Gelly, Sylvain},
  booktitle={ICML},
  year={2019}
}

@inproceedings{hu2022lora,
title={Lo{RA}: Low-Rank Adaptation of Large Language Models},
author={Edward J Hu and Yelong Shen and Phillip Wallis and Zeyuan Allen-Zhu and Yuanzhi Li and Shean Wang and Lu Wang and Weizhu Chen},
booktitle={ICLR},
year={2022}
}

@article{liang2024pointgst,
  title={Parameter-Efficient Fine-Tuning in Spectral Domain for Point Cloud Learning},
  author={Liang, Dingkang and Feng, Tianrui and Zhou, Xin and Zhang, Yumeng and Zou, Zhikang and Bai, Xiang},
  journal={TPAMI},
  year={2025}
}

@inproceedings{zha2025pma,
  title={PMA: Towards Parameter-Efficient Point Cloud Understanding via Point Mamba Adapter},
  author={Zha, Yaohua and Wang, Yanzi and Guo, Hang and Wang, Jinpeng and Dai, Tao and Chen, Bin and Ouyang, Zhihao and Yuerong, Xue and Chen, Ke and Xia, Shu-Tao},
  booktitle={CVPR},
  year={2025}
}

@inproceedings{han2025most,
  title={MoST: Efficient Monarch Sparse Tuning for 3D Representation Learning},
  author={Han, Xu and Tang, Yuan and Xu, Jinfeng and Li, Xianzhi},
  booktitle={CVPR},
  year={2025}
}

@inproceedings{ai2025gaprompt,
  author    = {Ai, Zixiang and Liu, Zichen and Lei, Yuanhang and Cui, Zhenyu and Zou, Xu and Zhou, Jiahuan},
  title     = {{GAP}rompt: Geometry-Aware Point Cloud Prompt for 3{D} Vision Model},
  booktitle = {ICML},
  year      = {2025}
}

@inproceedings{wu20153d,
  title={3D ShapeNets: A Deep Representation for Volumetric Shapes},
  author={Wu, Zhirong and Song, Shuran and Khosla, Aditya and Yu, Fisher and Zhang, Linguang and Tang, Xiaoou and Xiao, Jianxiong},
  booktitle={CVPR},
  year={2015}
}

@inproceedings{uy-scanobjectnn-iccv19,
      title = {Revisiting Point Cloud Classification: A New Benchmark Dataset and Classification Model on Real-World Data},
      author = {Mikaela Angelina Uy and Quang-Hieu Pham and Binh-Son Hua and Duc Thanh Nguyen and Sai-Kit Yeung},
      booktitle = {ICCV},
      year = {2019}
  }

@inproceedings{dongautoencoders,
  title={Autoencoders as Cross-Modal Teachers: Can Pretrained 2D Image Transformers Help 3D Representation Learning?},
  author={Dong, Runpei and Qi, Zekun and Zhang, Linfeng and Zhang, Junbo and Sun, Jianjian and Ge, Zheng and Yi, Li and Ma, Kaisheng},
  booktitle={ICLR},
  year = {2023}
}

@inproceedings{jia2022visual,
  title={Visual prompt tuning},
  author={Jia, Menglin and Tang, Luming and Chen, Bor-Chun and Cardie, Claire and Belongie, Serge and Hariharan, Bharath and Lim, Ser-Nam},
  booktitle={ECCV},
  year={2022},
  organization={Springer}
}

@inproceedings{liu2022masked,
  title={Masked discrimination for self-supervised learning on point clouds},
  author={Liu, Haotian and Cai, Mu and Lee, Yong Jae},
  booktitle={ECCV},
  year={2022},
  organization={Springer}
}

@inproceedings{vaswani2017attention,
  title={Attention is all you need},
  author={Vaswani, Ashish and Shazeer, Noam and Parmar, Niki and Uszkoreit, Jakob and Jones, Llion and Gomez, Aidan N and Kaiser, {\L}ukasz and Polosukhin, Illia},
booktitle={NeurIPS},
  year={2017}
}

@article{SUN2026112800,
title = {HyperPoint: Multimodal 3D foundation model in hyperbolic space},
author = {Yiding Sun and Haozhe Cheng and Chaoyi Lu and Zhengqiao Li and Minghong Wu and Huimin Lu and Jihua Zhu},
journal = {PR},
year = {2026}
}

@inproceedings{liu2019relation,
  title={Relation-shape convolutional neural network for point cloud analysis},
  author={Liu, Yongcheng and Fan, Bin and Xiang, Shiming and Pan, Chunhong},
  booktitle={CVPR},
  year={2019}
}

@article{shapenetpart,
  title={A scalable active framework for region annotation in 3D shape collections},
  author={ Yi, Li  and  Kim, Vladimir G.  and  Ceylan, Duygu  and  Shen, I Chao  and  Yan, Mengyan  and  Su, Hao  and  Lu, Cewu  and  Huang, Qixing  and  Sheffer, Alla  and  Guibas, Leonidas },
  journal={TOG},
  year={2016},
}

@inproceedings{bahri2025spectral,
  title={Spectral informed mamba for robust point cloud processing},
  author={Bahri, Ali and Yazdanpanah, Moslem and Noori, Mehrdad and Dastani, Sahar and Cheraghalikhani, Milad and Hakim, Gustavo Adolfo Vargas and Osowiechi, David and Beizaee, Farzad and Ben Ayed, Ismail and Desrosiers, Christian},
  booktitle={CVPR},
  year={2025}
}

@inproceedings{li2021prefix,
  title={Prefix-tuning: Optimizing continuous prompts for generation},
  author={Li, Xiang Lisa and Liang, Percy},
  booktitle={ACL},
  year={2021}
}

@inproceedings{jie2023revisiting,
  title={Revisiting the parameter efficiency of adapters from the perspective of precision redundancy},
  author={Jie, Shibo and Wang, Haoqing and Deng, Zhi-Hong},
  booktitle={ICCV},
  year={2023}
}

@article{sun2026align,
  title={Align then adapt: Rethinking parameter-efficient transfer learning in 4d perception},
  author={Sun, Yiding and Zhu, Jihua and Cheng, Haozhe and Luv, Chaoyi and Yang, Zhichuan and Chen, Lin and Wang, Yaonan},
  journal={TMM},
  year={2026}
}

@article{zhang2026pointcot,
  title={Pointcot: A multi-modal benchmark for explicit 3d geometric reasoning},
  author={Zhang, Dongxu and Sun, Yiding and Li, Pengcheng and Liu, Yumou and Lin, Hongqiang and Xu, Haoran and Mu, Xiaoxuan and Lin, Liang and Yan, Wenbiao and Yang, Ning and others},
  journal={arXiv preprint arXiv:2602.23945},
  year={2026}
}

@article{wang2026pointrft,
  title={PointRFT: Explicit Reinforcement Fine-tuning for Point Cloud Few-shot Learning},
  author={Wang, Yankai and Sun, Yiding and Wang, Qirui and Li, Pengbo and Lu, Chaoyi and Zhang, Dongxu},
  journal={arXiv preprint arXiv:2603.23957},
  year={2026}
}

@inproceedings{li2025pointdico,
  title={PointDico: Contrastive 3D Representation Learning Guided by Diffusion Models},
  author={Li, Pengbo and Sun, Yiding and Cheng, Haozhe},
  booktitle={IJCNN},
  year={2025}
}

@InProceedings{Armeni_2016_CVPR,
author = {Armeni, Iro and Sener, Ozan and Zamir, Amir R. and Jiang, Helen and Brilakis, Ioannis and Fischer, Martin and Savarese, Silvio},
title = {3D Semantic Parsing of Large-Scale Indoor Spaces},
booktitle = {CVPR},
year = {2016}
}

@inproceedings{ren2023spiking,
  title     = {Spiking PointNet: Spiking Neural Networks for Point Clouds},
  author    = {Ren, Dayong and Ma, Zhe and Chen, Yuanpei and Peng, Weihang and Liu, Xiaode and Zhang, Yuhan and Guo, Yufei},
  booktitle = {NeurIPS},
  year      = {2023}
}

@inproceedings{wu2025spiking,
  title     = {Spiking Point Transformer for Point Cloud Classification},
  author    = {Wu, Peixi and Chai, Bosong and Li, Hebei and Zheng, Menghua and Peng, Yansong and Wang, Zeyu and Nie, Xuan and Zhang, Yueyi and Sun, Xiaoyan},
  booktitle = {AAAI},
  year      = {2025}
}

@inproceedings{qiu2025efficient,
  title     = {Efficient 3D Recognition with Event-Driven Spike Sparse Convolution},
  author    = {Qiu, Xuerui and Yao, Man and Zhang, Jieyuan and Chou, Yuhong and Qiao, Ning and Zhou, Shibo and Xu, Bo and Li, Guoqi},
  booktitle = {AAAI},
  year      = {2025}
}

@inproceedings{wu2025efficient,
  title     = {Efficient Spiking Point Mamba for Point Cloud Analysis},
  author    = {Wu, Peixi and Chai, Bosong and Zheng, Menghua and Li, Wei and Hu, Zhangchi and Chen, Jie and Zhang, Zheyu and Li, Hebei and Sun, Xiaoyan},
  booktitle = {ICCV},
  year      = {2025}
}

@inproceedings{dang2026primary,
  title     = {Primary Visual Cortex Inspired Point Cloud Analysis Framework},
  author    = {Dang, Jisheng and Deng, Delin and Wang, Bimei and Wu, Jingze and Zhang, Hui and Li, Haijiang and Jiao, Jingmei and Pan, Dengyue and Xie, Mangang and Liu, Jizhao},
  booktitle = {AAAI},
  year      = {2026}
}

@inproceedings{he2026hybrid,
  title     = {3DSMT: A Hybrid Spiking Mamba-Transformer for Point Cloud Analysis},
  author    = {He, Yong and Wu, Qiaoyun and Mu, Chaoxu and Mian, Ajmal Saeed},
  booktitle = {ICLR},
  year      = {2026}
}

@inproceedings{wu2024point,
  title     = {Point-to-Spike Residual Learning for Energy-Efficient 3D Point Cloud Classification},
  author    = {Wu, Qiaoyun and Zhang, Quanxiao and Tan, Chunyu and Zhou, Yun and Sun, Changyin},
  booktitle = {AAAI},
  year      = {2024}
}

@inproceedings{lu2026spiking,
  title     = {Spiking Discrepancy Transformer for Point Cloud Analysis},
  author    = {Lu, Yijie and Pan, Zhiyi and Zhang, Renrui and Jia, Yanhao and Wang, Ronggang and Zhou, Zhaokun},
  booktitle = {ICLR},
  year      = {2026}
}

@article{qiu2025svl,
  title={SVL: Spike-based Vision-language Pretraining for Efficient 3D Open-world Understanding},
  author={Qiu, Xuerui and Wu, Peixi and Wen, Yaozhi and Gu, Shaowei and Pan, Yuqi and Luo, Xinhao and Xu, Bo and Li, Guoqi},
  journal={arXiv preprint arXiv:2505.17674},
  year={2025}
}

@article{pei2019towards,
  title     = {Towards Artificial General Intelligence with Hybrid Tianjic Chip Architecture},
  author    = {Pei, Jing and Deng, Lei and Song, Sen and Zhao, Mingguo and Zhang, Youhui and Wu, Shuang and Wang, Guanrui and Zou, Zhe and Wu, Zhenzhi and He, Wei},
  journal   = {Nature},
  year      = {2019}
}

@inproceedings{lan2023efficient,
  title={Efficient converted spiking neural network for 3d and 2d classification},
  author={Lan, Yuxiang and Zhang, Yachao and Ma, Xu and Qu, Yanyun and Fu, Yun},
  booktitle={ICCV},
  year={2023}
}

@article{fang2023spikingjelly,
  title={SpikingJelly: An Open-Source Machine Learning Infrastructure Platform for Spike-Based Intelligence},
  author={Fang, Wei and Chen, Yanqi and Ding, Jianhao and Yu, Zhaofei and Masquelier, Timoth{\'e}e and Chen, Ding and Huang, Liwei and Zhou, Huihui and Li, Guoqi and Tian, Yonghong},
  journal={Science Advances},
  year={2023}
}

@inproceedings{sedighi2024visual,
  title={Visual analysis of leaky integrate-and-fire spiking neuron models and circuits},
  author={Sedighi, Sara and Afrin, Farhana and Onyejegbu, Elonna and Cantley, Kurtis D},
  booktitle={MWSCAS},
  year={2024},
}

@INPROCEEDINGS{6757323,
  title={1.1 Computing's energy problem (and what we can do about it)}, 
  author={Horowitz, Mark},
  booktitle={ISSCC}, 
  year={2014}
}

@article{paszke2019pytorch,
  title={Pytorch: An imperative style, high-performance deep learning library},
  author={Paszke, Adam and Gross, Sam and Massa, Francisco and Lerer, Adam and Bradbury, James and Chanan, Gregory and Killeen, Trevor and Lin, Zeming and Gimelshein, Natalia and Antiga, Luca and others},
  journal={NeurIPS},
  year={2019}
}

@inproceedings{sun2026tri,
  title={Tri-Efficient Transfer Learning for Point Cloud Videos},
  author={Sun, Yiding and Zhang, Dongxu and Zhu, Jihua and Cheng, Haozhe and Li, Zhengqiao and Li, Pengcheng and Fang, Chaowei and Dong, Yonghao and Chen, Lin},
  booktitle={ECCV},
  year={2026}
}

@inproceedings{dai2017scannet,
  title={Scannet: Richly-annotated 3d reconstructions of indoor scenes},
  author={Dai, Angela and Chang, Angel X and Savva, Manolis and Halber, Maciej and Funkhouser, Thomas and Nie{\ss}ner, Matthias},
  booktitle={CVPR},
  year={2017}
}

@inproceedings{rozenberszki2022language,
  title={Language-Grounded Indoor 3D Semantic Segmentation in the Wild},
  author={Rozenberszki, David and Litany, Or and Dai, Angela},
  booktitle = {ECCV},
  year={2022}
}

@inproceedings{yeshwanth2023scannet++,
  title={Scannet++: A high-fidelity dataset of 3d indoor scenes},
  author={Yeshwanth, Chandan and Liu, Yueh-Cheng and Nie{\ss}ner, Matthias and Dai, Angela},
  booktitle={ICCV},
  year={2023}
}

@inproceedings{caesar2020nuscenes,
  title={nuscenes: A multimodal dataset for autonomous driving},
  author={Caesar, Holger and Bankiti, Varun and Lang, Alex H and Vora, Sourabh and Liong, Venice Erin and Xu, Qiang and Krishnan, Anush and Pan, Yu and Baldan, Giancarlo and Beijbom, Oscar},
  booktitle={CVPR},
  year={2020}
}

@inproceedings{wu2025sonata,
  title={Sonata: Self-supervised learning of reliable point representations},
  author={Wu, Xiaoyang and DeTone, Daniel and Frost, Duncan and Shen, Tianwei and Xie, Chris and Yang, Nan and Engel, Jakob and Newcombe, Richard and Zhao, Hengshuang and Straub, Julian},
  booktitle={CVPR},
  year={2025}
}

@InProceedings{Liu_2026_CVPR,
    author    = {Liu, Siyuan and Zheng, Chaoqun and Zhou, Xin and Feng, Tianrui and Liang, Dingkang and Bai, Xiang},
    title     = {PointTPA: Dynamic Network Parameter Adaptation for 3D Scene Understanding},
    booktitle = {CVPR},
    year      = {2026}
}

\section{Biography Section}
\vspace{-1.1cm}
\begin{IEEEbiographynophoto}{Zihao Guo}
received the B.E. degree in computer science and technology from Fuzhou University, Fuzhou, China, in 2025. He is currently working toward the M.E. degree with the School of Software, Xi'an Jiaotong University, Xi'an, China. His research interests include 3D point cloud representation learning and spiking neural networks.
\end{IEEEbiographynophoto}
\vspace{-1.1cm}
\begin{IEEEbiographynophoto}{Jihua Zhu}
(Senior Member, IEEE) received the B.E. degree in automation from Central South University, China, and the Ph.D. degree in pattern recognition and intelligence systems from Xi'an Jiaotong University, China, in 2004 and 2011, respectively. Currently, he is a professor in the School of Software at Xi'an Jiaotong University. His research interests include computer vision and machine learning.
\end{IEEEbiographynophoto}
\vspace{-1.1cm}
\begin{IEEEbiographynophoto}{Yiding Sun}
received the B.E. degree in Software Engineering at the School of Computer Science and Engineering, Xi'an Technological University. He is currently pursuing his M.E. degree in the School of Software at Xi'an Jiaotong University. His research interests include 3D point cloud analysis.
\end{IEEEbiographynophoto}
\vspace{-1.1cm}
\begin{IEEEbiographynophoto}{Dongxu Zhang}
received the B.E. degree in computer science and technology from Taiyuan University of Technology, Taiyuan, China, in 2024. He is currently working toward the M.E. degree with the School of Software, Xi'an Jiaotong University, Xi'an, China. His research interests include point cloud registration, large language models, and computer vision.
\end{IEEEbiographynophoto}
\vspace{-1.1cm}
\begin{IEEEbiographynophoto}{Hainan Luo}
received the M.E. degree in Computer Technology from Anhui Polytechnic University, Wuhu, China, in 2024. He is currently the Deputy General Manager of Wuhu HIT Robot Technology Research Institute Co., Ltd. His research interests mainly include artificial intelligence and its applications. He has published several academic papers, including one paper in a CAS Zone 1 journal and one paper in a Chinese core journal.
\end{IEEEbiographynophoto}
\vspace{-1.1cm}
\begin{IEEEbiographynophoto}{Kai Tang}
received the B.Eng. degree in communication engineering from the University of Electronic Science and Technology of China, Chengdu, China, in 2025. He is currently pursuing the M.Comp. degree in applied artificial intelligence at Nanyang Technological University, Singapore, and is a visiting student with the School of Computer Science, Peking University, Beijing, China. His research interests include world models, vision-language-action models, reinforcement learning, and trustworthy embodied AI.
\end{IEEEbiographynophoto}
\vspace{-1.1cm}
\begin{IEEEbiographynophoto}{Shuaikang Zhu}
is currently working toward the M.E. degree in software engineering with Xi'an Jiaotong University, Xi'an, China. He has served as the technical leader of multiple industrial intelligent projects. His research interests include human pose estimation and human behavior recognition.
\end{IEEEbiographynophoto}
\vspace{-1.1cm}
\begin{IEEEbiographynophoto}{Lin Chen}
received the Ph.D. degree in Control Science and Engineering from Hunan University, Changsha, China, in 2025, the M.S. degree in computer technology from the University of Chinese Academy of Sciences, Beijing, China, in 2021, and the B.S. degree in Electronic Information Engineering from Northwest Normal University, Lanzhou, China, in 2018. He is currently an Assistant Professor with the School of Software Engineering, Xi'an Jiaotong University, Xi'an, China. His current research interests include multi-robot systems, deep reinforcement learning, and visual navigation.
\end{IEEEbiographynophoto}
\vspace{-1.1cm}
\begin{IEEEbiographynophoto}{Jian Liu}
received the Ph.D. degree in 2025 from the National Engineering Research Center of Robot Visual Perception and Control Technology, Hunan University, Changsha, China. He is currently a Research Fellow at the National University of Singapore. From 2023 to 2024, he was a visiting Ph.D. student with the Department of Computer Science, The University of Western Australia. His current research interests include robotic manipulation, 3D machine vision, and object pose estimation. He has served as a reviewer for more than 30 journals and conferences.
\end{IEEEbiographynophoto}
\vspace{-1.1cm}
\begin{IEEEbiographynophoto}{Ajmal Mian}
is a Professor of Computer Science at The University of Western Australia, Perth, WA, Australia. His research interests include computer vision, machine learning, and artificial intelligence. He has received the Vice-Chancellor's Mid-Career Research Award and the IAPR Best Scientific Paper Award. He was an ARC Future Fellow, an IAPR Fellow, an ACM Distinguished Speaker, and the President of the Australian Pattern Recognition Society. He frequently serves as an Associate Editor or Area Chair for top-tier venues, including IEEE TPAMI and CVPR.
\end{IEEEbiographynophoto}

\vfill

\end{document}